\documentclass[preprint,review,12pt]{elsarticle}

\usepackage{amsmath}
\usepackage{amssymb}
\usepackage{booktabs}
\usepackage{lineno}
\usepackage{algorithm}
\usepackage[noend]{algpseudocode}
\usepackage{geometry}
\usepackage{svg}
\usepackage{graphicx}
\usepackage{listings}
\usepackage{tcolorbox}
\usepackage{comment}

\usepackage{graphicx}
\usepackage{subcaption}
\usepackage{caption} 

\usepackage{multirow}
\usepackage{xcolor}
\usepackage{longtable}
\usepackage{multirow}
\usepackage{array}
\usepackage{tabularx}
\usepackage{rotating}
\usepackage{arydshln}

\newcommand{\revision}[1]{#1}

\definecolor{roomcolor}{HTML}{bfdae5}     
\definecolor{staircolor}{HTML}{d6a33b}    
\definecolor{doorcolor}{HTML}{765241}     

\usepackage{pgfplots}

\definecolor{erdline}{HTML}{8C8C8C}
\definecolor{erdtype}{HTML}{6B6B6B}
\definecolor{erdkey}{HTML}{222222}

\usepackage{setspace}

\usepackage{url}

\usepackage{enumitem}
\definecolor{codebg}{RGB}{245,247,250}      
\definecolor{framegray}{RGB}{180,185,190}   
\definecolor{textgray}{RGB}{30,30,30}       
\definecolor{keywordcolor}{RGB}{0,102,204} 
\definecolor{commentcolor}{RGB}{128,128,128} 
\definecolor{stringcolor}{RGB}{153,0,0}    

\lstdefinestyle{querylog}{
    basicstyle=\ttfamily\footnotesize\color{textgray},
    keywordstyle=\color{keywordcolor}\bfseries,
    commentstyle=\color{commentcolor}\itshape,
    stringstyle=\color{stringcolor},
    breaklines=true,
    frame=single,
    backgroundcolor=\color{codebg},
    rulecolor=\color{framegray},
    framesep=12pt,
    xleftmargin=15pt,
    xrightmargin=15pt,
    framexleftmargin=0pt,
    framexrightmargin=0pt,
    aboveskip=15pt,
    belowskip=15pt,
    columns=fullflexible,
    keepspaces=true,
    showstringspaces=false,
    rulecolor=\color{framegray},
    captionpos=b
}

\journal{Elsevier}
\pgfplotsset{compat=1.18}

\begin{document}
\begin{frontmatter}



\title{IFC-MCQ: Automatically Generating Multiple-Choice Benchmarks for Evaluating LLM Agents on IFC Data Retrieval}


\author[grid]{Rabindra Lamsal\corref{cor1}}
\ead{r.lamsal@unsw.edu.au}

\author[grid]{Sisi Zlatanova}
\ead{s.zlatanova@unsw.edu.au}

\author[civil]{Xuesong Shen}
\ead{x.shen@unsw.edu.au}

\affiliation[grid]{organization={GRID Lab, School of Built Environment, The University of New South Wales},
            city={Sydney},
            country={Australia}}

\affiliation[civil]{organization={School of Civil and Environmental Engineering, The University of New South Wales},
            city={Sydney},
            country={Australia}}

\cortext[cor1]{Corresponding author}
\begin{abstract}
Industry Foundation Classes (IFC), the standard for building information exchange in architecture, engineering, and construction (AEC), is increasingly queried through large language models (LLMs) using natural language. Evaluating these systems at scale is difficult because free-text answers require costly human or LLM-based review. We propose IFC-MCQ, a method that automatically generates multiple-choice benchmarks from IFC files for exact option-set grading. The taxonomy covers 15 categories and 36 subtypes across four groups, with answer keys derived from the file and distractors constructed from file-derived values or predefined rules. We applied IFC-MCQ to IFC models of varying sizes and modelling characteristics to assess generation coverage and evaluated two agentic configurations on selected models using tool- and code-based retrieval. The evaluation exposed category-specific failure modes and showed that performance can vary with the source IFC model and retrieval interface. IFC-MCQ enables scalable benchmarking of LLM-based IFC retrieval in AEC workflows.
\end{abstract}




\end{frontmatter}



\section{Introduction}
\label{sec:intro}
Industry Foundation Classes (IFC) standard is an open, standardized data model widely used for exchanging building information across the architecture, engineering, construction (AEC) sector, and a range of building management and exploration applications \citep{eastman2011bim,wang2019integration,liu2021indoor,alattas2021mapping,aleksandrov2024bim}. An IFC file encodes building elements, their attributes, property and quantity sets, materials, classification references, spatial structure and geometry in a single semantic model \citep{ifc}. Querying IFC data is a common task, and its user base is expanding well beyond architects and engineers \citep{eastman2011bim}. For instance, facility managers may query an IFC model to identify the number of rooms per storey in order to allocate workforce, and energy consultants might query the thermal transmittance of walls to check for compliance. However, querying an IFC model is a demanding task that requires either custom scripting against an IFC toolkit or detailed knowledge of the complex and deeply-nested IFC schema. 

Ongoing progress in large language models (LLMs) \citep{zhao2023survey,achiam2023gpt} has made natural language querying of building information a more accessible option \citep{pan2026llm}, and a growing body of work has proposed LLM-based natural language interfaces for IFC models \citep{iranmanesh2025llm,ibba2026ask,avgoren2025enhancing,lamsal2026ifcllm,ifcbenchv2}, allowing non-experts to query such models in natural language. However, such an interface is only useful if its answers are correct. An LLM that confidently presents a plausible but incorrect fire rating, a quantity in the wrong unit, or a wall assigned to the wrong storey can lead to incorrect downstream decisions. Therefore, rigorous evaluation of these systems is critical, but this remains a challenge. IFC data retrieval tasks are inherently open-ended \citep{ifcbenchv1,lamsal2026ifcllm}, and grading open-ended question-answering reliably at scale is expensive \citep{balepur2025these}. Grading a free-text response requires either costly human review or a second LLM whose own judgements carry additional, hard-to-quantify uncertainty, and relying on an LLM for review only adds to the overall cost of inference. An automated approach to grading is therefore needed.

Multiple-choice questions (MCQs) provide a practical alternative \citep{robinson2022leveraging,balepur2025these}. Formally, following the multiple-choice prompting (MCP) strategy of \citet{robinson2022leveraging}, an MCQ can be defined as a tuple $(q, A, A^*)$, where $q$ is the question in natural language (question stem), $A=\{a_1, a_2, \dots, a_n\}$ is a finite set of options (answer key(s) + distractors), and $A^* \subseteq A$ is the set of correct answer(s). A response is scored as correct only if the selected answer(s) $\hat{A} \subseteq A$ exactly match $A^*$. For example, given $q$: ``How many storeys does the building model contain?'' with $A=\{1, 2, 3, 4, 5\}$ and $A^*=\{3\}$, an LLM is correct only if it selects ``3''. Outside the IFC domain, MCQs are widely used for evaluating foundational LLMs \citep{bommasani2021opportunities}, including benchmarks such as ARC \citep{clark2018think}, MMLU \citep{hendrycks2020measuring}, and GPQA \citep{rein2023gpqa}. However, to the best of our knowledge, MCQ-based benchmarks have not been established for natural language querying of IFC models. Developing such a benchmark requires three capabilities: (i) a taxonomy of question types that are meaningful for IFC querying and admit reliable ground truth; (ii) a mechanism for generating questions directly from IFC models; and (iii) a separate procedure for verifying the correctness of generated items.

\revision{In this work, we address these requirements as follows. We introduce IFC-MCQ, a method for automatically generating MCQ benchmarks from IFC files. First, it defines a taxonomy of 15 question categories (further divided into 36 subtypes) organized into four groups: Direct retrieval, Relational, Quantitative, and Geometric. The taxonomy focuses on questions for which the IFC file can supply reliable ground truth and is meant to be extensible. Second, IFC-MCQ generates benchmark items directly from the content of the input IFC file. Each category generator uses the information available in the file to construct question stems, answer key(s), and distractors designed to represent IFC-specific retrieval errors, such as values taken from other entities of the same type, quantities expressed in an incorrect unit, or plausible values for properties that are absent from the queried entity. Third, the answer to each candidate item is re-derived through a separate verification path that does not receive the stored answer key. Candidates for which the generated and re-derived answers disagree are discarded.}

\revision{To demonstrate the method, we first apply IFC-MCQ to ten IFC models of varying sizes and modelling characteristics to examine benchmark coverage and verification across heterogeneous files. We then evaluate the generated benchmarks through a three-stage agentic experiment. Stage 1 evaluates 17 proprietary and open-weight LLMs using a tool agent on two commonly used reference IFC models. Stages 2 and 3 use a selected set of LLMs across the two reference models and four additional IFC models, while introducing a second agent, i.e., a code agent, to examine the effects of source-model characteristics and agent design, respectively.}

The remainder of the paper is organized as follows. Section \ref{sec:literature} reviews the relevant literature; Section \ref{sec:method} describes the method; Section \ref{sec:method-eval} discusses the implementation and the evaluation setup; Section \ref{sec:results} presents the results; Section \ref{sec:discussion} provides discussion; and Section \ref{sec:conclusion} concludes the paper.

\section{Literature Review}
\label{sec:literature}

This section reviews existing approaches to natural language querying of IFC models, and the benchmarks used to evaluate them. Section~\ref{sec:lit-systems} reviews querying systems, from early semantic-parsing to recent LLM-based approaches, and Section~\ref{sec:lit-benchmarks} covers benchmarks built specifically for evaluating such systems. Section~\ref{sec:lit-gap} then identifies the gap that motivates IFC-MCQ.

\subsection{Natural language querying systems for IFC} 
\label{sec:lit-systems}

Earlier work treats the natural language interface to IFC models as a semantic parsing problem: translating a natural language question into a formal query over the model's structured representation. \citet{lin2016natural}, for example, extract keywords from a user's input and map these to IFC entities or attributes. \citet{shin2021bimasr} develop BIMASR, a system that accepts voice input and analyzes it for syntax and semantics. \citet{yin2023two} propose a two-stage Text-to-BIMQL approach that combines graph neural networks with a domain ontology to convert natural language input into executable queries. These studies, however, are largely limited to extracting entities and relationships from a natural language question, and fall short once the question requires more complex reasoning \citep{yin2023two}.

Recent work follows a similar overall design, transforming an IFC model into a structured representation and using a language model to write the query, but replaces hand-crafted rules with retrieval-augmented generation (RAG) \citep{lewis2020retrieval, gao2023retrieval}, most often over a graph representation of IFC data. ASK-BIM \citep{ibba2026ask} converts an IFC model into a knowledge graph and uses an LLM to generate SPARQL queries for data retrieval and, finally, to compose a natural language answer. It is evaluated on a single multi-storey building against a manually created set of 28 questions gathered from domain professionals and graded manually. \citet{iranmanesh2025llm} transform an IFC model into a Neo4j property graph and instruct GPT-4o to generate Cypher \citep{francis2018cypher}, evaluating the system on one building model against a small, author-created set of 60 questions used as ground truth. \citet{avgoren2025enhancing} similarly build a knowledge graph that explicitly encodes spatial and relational dependencies between building elements, and embed it in an LLM-based multi-agent system for IFC data retrieval. In particular, the authors reduce the full set of IFC relationships to a smaller number of core types (e.g., hasStorey, hasSpace) to lower the reasoning complexity for the LLM. Their system is evaluated on 135 manually created questions across three categories: Single entity inquiry, Multiple entities, and Mathematical operations. BIMConverse \citep{pacheco2024bimconverse} applies a similar graph-and-Cypher strategy with GPT-4o, though the authors describe its behaviour qualitatively through example queries rather than a formal benchmark or quantitative evaluation. BIMCoder \citep{liu2025bimcoder} takes a different route: rather than retrieval-augmented generation, it uses pretrained language models to translate natural language directly into the structured query statements used by professional BIM software, over a purpose-built dataset of 1,680 query pairs. Because the target is a formal query, grading is fully automated, but this measures the correctness of the query rather than of the answer a user ultimately receives.

More recently, hybrid IFC data representations and native IFC file exploration have been implemented within LLM-based agentic systems. IfcLLM \citep{lamsal2026ifcllm} transforms an IFC model into complementary relational and graph databases and uses an open-weight LLM to route each query to the more suitable backend. The authors evaluated the system on three openly available models against their own set of 30 query scenarios, with correctness checked manually against expected answers. \citet{ifcbenchv2} propose an adaptive exploration approach in which an LLM-based agent accesses an IFC model programmatically, iteratively generating and executing Python code that discovers the model's structure at run time. Their approach is evaluated on a benchmark of 1,027 question-answer pairs, with grading done by an LLM.

\subsection{Benchmarks for IFC querying} 
\label{sec:lit-benchmarks}
Most of the systems reviewed above are evaluated on small, author-curated question sets. More recently, dedicated benchmarks have begun to emerge. IFC-Bench \citep{ifcbenchv1, ifcbenchv2} is an openly available benchmark for evaluating IFC data retrieval systems, released in two versions. IFC-Bench v1 contains 105 question-answer pairs for two IFC models; IFC-Bench v2 extends this to 1,027 pairs across 37 IFC models, organized into four categories: (i) direct retrieval (explicitly stored properties) (ii) computational aggregation (counting or summation); (iii) geometric/spatial computation (geometric properties or spatial relationships) and (iv) incomplete information (questions requiring assumptions or acknowledging missing information). In both versions, answers are free text, so grading relies on an LLM. IFC-QA \citep{volkova2026ifc} is a benchmark of 99 natural language questions, paired with ground-truth extraction scripts, organized into five categories: (i) factual (e.g., number of doors), (ii) attribute aggregation (e.g., total floor area), (iii) relational aggregation (e.g., total area of the curtain walls), (iv) multi-hop (e.g., rooms without windows), and (v) constraint (e.g., spaces >50 m$^2$). IFCMemoryBench \citep{duifcmemorybench} extends the setting to multi-session interaction, with 143 human-validated tasks derived from the incomplete-information questions in \citep{ifcbenchv2}.

\subsection{Summary and gap}
\label{sec:lit-gap}

On the systems side, most quantitatively evaluated approaches use one or a few IFC models, small author-curated question sets, and manual grading \citep{iranmanesh2025llm,ibba2026ask,avgoren2025enhancing,lamsal2026ifcllm}. Where evaluation is automated, the target is typically a formal query with executable ground truth, rather than the answer ultimately presented to a user \citep{yin2023two,liu2025bimcoder,volkova2026ifc}. The code-executing agent evaluated by \citet{ifcbenchv2} uses a shared benchmark but relies on an LLM judge for grading. On the benchmark side, datasets built around free-text answers require either human review or an LLM whose own judgements may carry uncertainty \citep{ifcbenchv1,ifcbenchv2,duifcmemorybench}, while the benchmark with fully automatic grading evaluates the query pipeline rather than the reliability of the answer itself \citep{volkova2026ifc}. \revision{Consequently, a gap remains for a benchmark-generation method that can produce questions for new IFC files, verify the generated answer keys, and grade agent responses deterministically at the answer level.}

\revision{IFC-MCQ addresses this gap by combining automatic item generation, category-specific distractor construction, separate answer re-derivation, and exact option-set grading. Unlike the fixed question sets used in earlier benchmarks, IFC-MCQ can generate a new benchmark for a parseable IFC model when the file contains the facts and distinct answer alternatives required by the relevant categories. The method therefore supports evaluation across different source models while making category coverage explicitly dependent on the information represented in each file. Table~\ref{tab:summary} compares this scope with existing IFC querying systems and benchmarks.}

\begin{table}
\centering
\footnotesize
\caption{Natural language (NL) querying systems for IFC and related benchmarks.}
\label{tab:summary}
\begin{tabular}{@{}p{4cm} p{4cm} p{4cm} p{3cm}@{}}
\toprule
\textbf{Work} &
\textbf{What it does} &
\textbf{Evaluation set; grading} &
\textbf{Evaluation size} \\
\midrule

\multicolumn{4}{@{}l}{\textit{Systems}} \\
\addlinespace[2pt]

\citet{yin2023two}
& NL to BIMQL via GNN and ontology
& Study-specific; automated query matching
& Fixed query types \\

\citet{ibba2026ask}
& NL to SPARQL over an IFC knowledge graph
& Study-specific; manual
& 28 questions \\

\citet{iranmanesh2025llm}
& NL to Cypher over Neo4j
& Study-specific; manual
& 60 questions \\

\citet{avgoren2025enhancing}
& NL to Cypher over Neo4j with reduced IFC relationships
& Study-specific; manual
& 135 questions \\

\citet{pacheco2024bimconverse}
& NL to Cypher over Neo4j
& Qualitative examples; no formal grading
& Not reported \\

\citet{lamsal2026ifcllm}
& NL to SQL and/or Cypher over SQLite and Neo4j
& Study-specific; manual
& 30 scenarios \\

\citet{liu2025bimcoder}
& NL to structured BIM query statements
& Study-specific; automated query matching
& 1{,}680 pairs \\

\citet{ifcbenchv2}
& Code-executing agent over IFC
& Benchmark; LLM-judged
& 1{,}027 questions \\

\midrule
\multicolumn{4}{@{}l}{\textit{Benchmarks}} \\
\addlinespace[2pt]

\citet{ifcbenchv1,ifcbenchv2}
& Human-authored QA benchmark with four categories
& Benchmark; LLM-judged
& 105 (v1); 1{,}027 (v2) \\

\citet{volkova2026ifc}
& Human-authored QA benchmark with five categories
& Benchmark; automated query matching
& 99 questions \\

\citet{duifcmemorybench}
& Benchmark derived from the \textit{incomplete-information} category in \cite{ifcbenchv2}
& Benchmark; human validation and LLM-judged
& 143 tasks \\

\textbf{IFC-MCQ} (this work)
& Automatically generated MCQ benchmark with 15 categories and 36 subtypes
& Benchmark; exact option-set grading
& Varies by IFC model and per-category generation cap \\

\bottomrule
\end{tabular}
\end{table}

\section{Method}
\label{sec:method}

IFC-MCQ constructs an MCQ benchmark in which each correct option-set is derived from the underlying IFC file. Distractors are drawn from other values in the same file wherever possible and are otherwise produced using predefined, category-specific transformations. Benchmark generation is controlled by the input IFC file, the implementation version, and a fixed random seed. The following subsections describe the method in detail: first the question taxonomy, then the IFC-MCQ pipeline.

\subsection{Question taxonomy}
\label{sec:method-taxonomy}

We define a question taxonomy of 15 categories and 36 subtypes, organized into four higher-level groups, as shown in Table~\ref{tab:taxonomy}. \revision{The four groups were inspired by the question categories of IFC-Bench v2 \citep{ifcbenchv2} and IFC-QA \citep{volkova2026ifc}, which are broad and defined largely by the type of answer sought. Such broad categories make it difficult to identify which type of retrieval a system handles poorly. We therefore divided each group by the IFC mechanism a question has to read, which gives the 15 categories, and divided categories further into subtypes that differ in the probe or the direction of traversal (e.g., opening to host, filler to host, and host to fillers).} Group~1 (\textit{Direct retrieval}) asks for a fact attached to a single entity: its IFC class, an attribute, a named property, a boolean claim, or whether a named property is populated at all. Group~2 (\textit{Relational}) requires traversing IFC relationships, such as a type definition, spatial containment or aggregation, a material association, an opening and filling chain, or a physical space boundary. Group~3 (\textit{Quantitative}) covers counting, quantity retrieval, aggregation, and comparison. Group~4 (\textit{Geometric}) covers questions that require reasoning over the geometry of one or more elements. This grouping lets us report evaluation metrics by the kind of reasoning a question requires, and it is not meant to be exhaustive: further groups, categories, or subtypes can be added as the benchmark is extended beyond what the current version covers. 

\begin{table}
\centering
\caption{IFC-MCQ question taxonomy. Groups organize categories by interaction type. Some categories further divide into subtypes (refer to Table~\ref{tab:subtypes}).}
\label{tab:taxonomy}
\footnotesize
\begin{tabular}{@{}lll@{}}
\toprule
\textbf{Group} & \textbf{Cat.} & \textbf{Name} \\
\midrule
\multirow{5}{*}{(1) Direct retrieval} & CLS & Class identification \\
 & ATT & Attribute retrieval \\
 & PRP & Property retrieval \\
 & BLN & Boolean property verification \\
 & EXS & Existence / absence \\
\midrule
\multirow{5}{*}{(2) Relational} & TYP & Type object resolution \\
 & SPA & Spatial structure \\
 & MAT & Material association \\
 & OPN & Openings and fillings \\
 & BND & Space boundaries \\
\midrule
\multirow{4}{*}{(3) Quantitative} & CNT & Counting \\
 & QTY & Quantity retrieval \\
 & AGG & Aggregation \\
 & CMP & Comparison / extremum \\
\midrule
(4) Geometric & GEO & Geometry \\
\bottomrule
\end{tabular}
\end{table}

Each benchmark item also takes one of three question types: single-choice items contain one correct option among several distractors, boolean items require a Yes/No or True/False response, and multi-choice items require identifying all correct options. Table~\ref{tab:subtypes} gives an overview of the subtypes, question types, and probes (input-output) considered in this work.

\begin{table}
\centering
\footnotesize
\setlength{\tabcolsep}{4pt}
\renewcommand{\arraystretch}{1.05}
\caption{Categories and their subtypes considered in this work. $^\dagger$S (single-choice), M (multi-choice), B (boolean.)}
\label{tab:subtypes}
\begin{tabular}{@{}lll p{9.5cm}@{}}
\toprule
\textbf{Cat.} & \textbf{Subtype} & \textbf{Type$^\dagger$} & \textbf{Probe (input $\to$ output)} \\
\midrule
\multicolumn{2}{l}{\textbf{\underline{Direct retrieval}}}  \\
CLS & class by GUID     & S & entity GUID $\to$ IFC class. \\
ATT & attribute value   & S & entity, attribute $\to$ stored value. \\
PRP & property value    & S & entity, pset, property $\to$ stored value. \\
BLN & boolean check     & B & entity, claimed Boolean $\to$ whether it matches. \\
EXS & property existence & B & entity, class-common property $\to$ whether populated. \\
    & value or null     & S & As above $\to$ stored value, or \textit{Not specified in the file}. \\
\midrule
\multicolumn{2}{l}{\textbf{\underline{Relational}}}  \\
TYP & type object       & S & element $\to$ its type object. \\
SPA & element's storey   & S & element $\to$ storey that contains it. \\
    & space's storey     & S & space $\to$ its storey. \\
    & storey's building  & S & storey $\to$ building that aggregates it. \\
MAT & material name     & S & element with one material $\to$ material name. \\
    & material set      & M & element with a material set $\to$ every material name. \\
OPN & opening to host   & S & opening $\to$ host element it voids. \\
    & filler to host    & S & filler $\to$ host element it fills. \\
    & host to fillers   & M & host element $\to$ all its fillers. \\
BND & element-to-spaces & M & element $\to$ every space it bounds. \\
    & boundary check    & B & space, element $\to$ whether a physical boundary relates them. \\
    & space-to-elements & M & space $\to$ every element that bounds it. \\
\midrule
\multicolumn{2}{l}{\textbf{\underline{Quantitative}}}  \\
CNT & file count        & S & IFC class $\to$ entity count in the file (subclasses incl.). \\
    & storey count      & S & class, storey $\to$ count on that storey. \\
    & type count        & S & type object $\to$ number of occurrences. \\
    & opening count     & S & host element $\to$ number of openings cut into it. \\
QTY & area / volume / length & S & entity, base quantity $\to$ value in m\textsuperscript{2} / m\textsuperscript{3} / m. \\
    & net vs.\ gross    & S & a \textit{Net}/\textit{Gross} pair $\to$ the larger of the two. \\
AGG & property sum / mean & S & class, numeric property $\to$ sum / mean of stored values. \\
    & quantity sum / mean & S & class, base quantity $\to$ SI sum / mean of stored values. \\
CMP & pairwise compare  & S & two entities, shared metric $\to$ the larger one. \\
    & extremum (largest / smallest) & S & two entities, shared metric $\to$ the entity with the largest / smallest value. \\
\midrule
\multicolumn{2}{l}{\textbf{\underline{Geometric}}}  \\
GEO & point in box      & B & element, world point (m) $\to$ whether the point lies inside element's AABB. \\
    & base elevation    & S & two elements $\to$ the one with higher AABB min $Z$. \\
    & extent compare    & S & two elements, an axis $\to$ larger extent along the axis. \\
\bottomrule
\end{tabular}
\end{table}

\paragraph{Scope} IFC-MCQ is currently scoped to data retrievable from the IFC schema, together with a small set of lightweight and deterministic computations derived directly from that data. This scope places a ceiling on the types of questions the benchmark can ask, particularly in Group~4. Advanced geometric reasoning typically requires boundary representation parsing or mesh-based operations that introduce substantially more geometric processing beyond simple IFC data retrieval. However, excluding geometric reasoning altogether would omit an important class of queries commonly used in BIM practice. We therefore retain Group~4 as a deliberately limited form of geometric reasoning, based on axis-aligned bounding boxes (AABBs). AABBs provide a simple and computationally inexpensive geometric abstraction that supports extent comparison, base elevation, and point-in-box queries without requiring solid-level geometric operations. Operations on solid geometry, such as boolean intersection, point-in-solid tests, or volume computations, are left for future work.

\subsection{IFC-MCQ Pipeline}
\label{sec:method-pipeline}

\begin{figure}
    \centering
    \vspace*{-2cm}
    \hspace*{-0.15\linewidth}
    \includegraphics[width=1.3\linewidth]{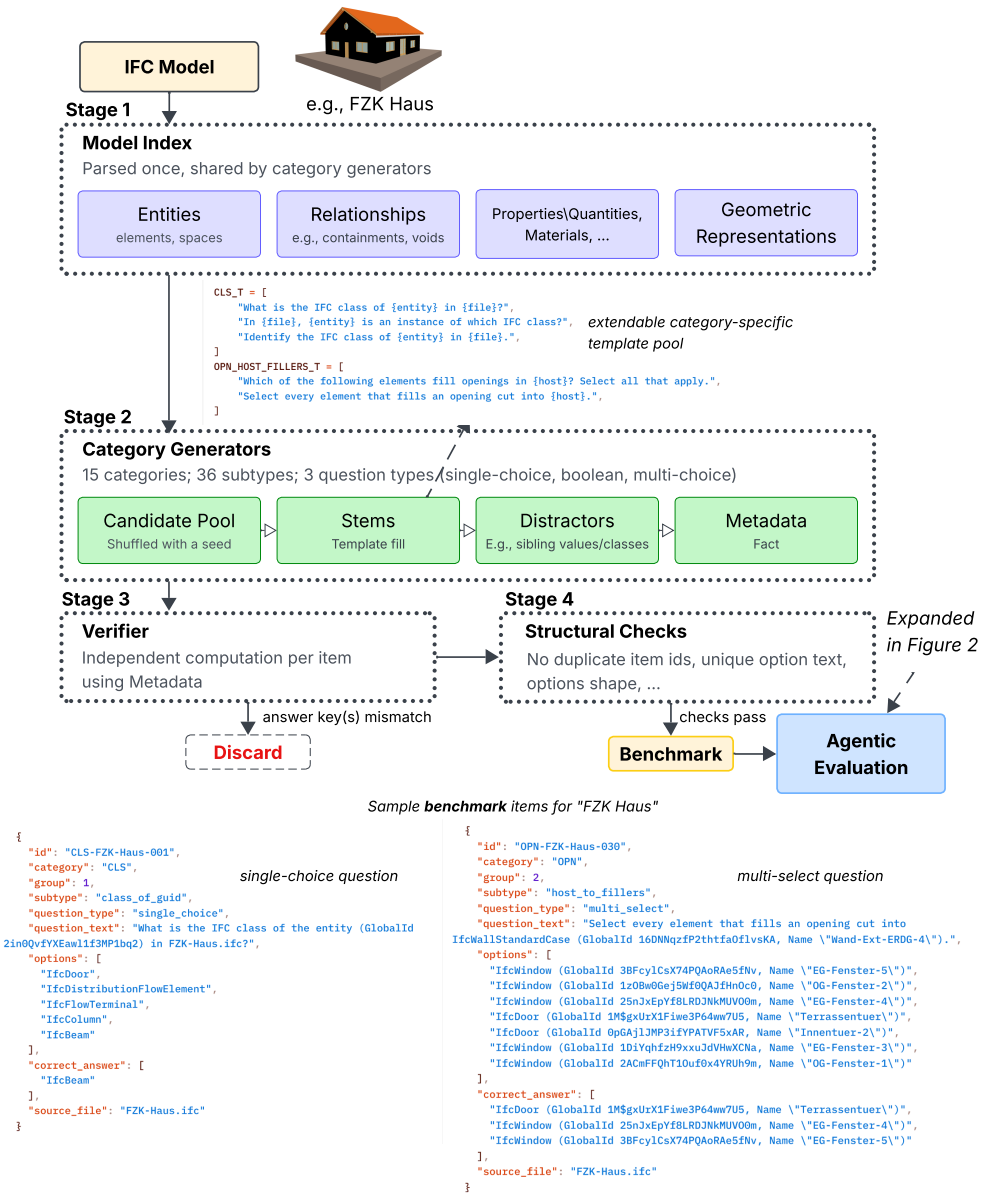}
    \caption{An overview of the IFC-MCQ pipeline.}
    \label{fig:method}
\end{figure}

Given an IFC file, the pipeline processes it in four stages, as illustrated in Figure \ref{fig:method}. The first stage builds a \textit{model index}: an inventory of the building's elements, together with their properties, relationships, and geometries, that determines which question categories are supported by the file. The second stage applies \textit{category generators}, which use the model index to produce candidate items for each category. The third stage employs \textit{verifiers} that separately re-derive the answer to each candidate item to confirm its correctness. Finally, the fourth stage performs \textit{structural checks} on the verified items, and only candidates that pass both answer verification and the structural checks are retained in the benchmark. Each benchmark item (refer to Figure \ref{fig:method} for sample items) includes the following: identifier, category, group, subtype, question type, question text (stem), options, correct answer, and source file.

\subsubsection{Model Index and Category Generators}
\label{sec:method-index-generation}

Building the model index is the first stage of the pipeline. Each IFC file is parsed into an in-memory model, on top of which a set of lookups is built for the generators to draw from. These lookups define the scope of the benchmark: a question is constructed around a fact that appears in the index or can be derived from it. Because further lookups can be added to broaden the range of supported question categories, this scope is easily extended. The discussion of the model index that follows describes the scope of the current version of the pipeline, not an upper bound on what the benchmark could eventually cover.

The index first collects the entities around which questions are built. It keeps all \textit{IfcElement} instances, along with the storeys, spaces, buildings, and sites in the file. Elements and spaces are grouped by their class. Next, the index reads the relationships that connect these entities. Spatial containment comes from \textit{IfcRelContainedInSpatialStructure}. Space-to-storey and storey-to-building links come from \textit{IfcRelAggregates}. Relationships between hosts, openings and fillers come from \textit{IfcRelVoidsElement} and \textit{IfcRelFillsElement}. Space and element pairs come from \textit{IfcRelSpaceBoundary}. Finally, each element is matched to its type object, and the reverse map from a type to its occurrences is stored.

The index also reads the data attached to each entity. Property sets are read as name-value pairs, and quantity sets are read for length, area, volume, count, and weight. Materials are resolved through \textit{IfcRelAssociatesMaterial}, covering a single material, a material list, and a layer set. A single length scale factor is taken from the project units, and values stay in model units, so a generator can scale length, area, and volume to SI units whenever required. Finally, the AABB of each building element is extracted from its triangulated mesh and stored.

\textit{(i) Candidate Pool.} With the model index built, the second stage turns its contents into candidate items. Every category generator follows the same basic procedure, whatever category it targets: it reads the part of the index relevant to its category and builds a pool of facts, where each fact pairs an entity, or a group of entities, with the datum the question will ask about, i.e., an attribute, a property, a spatial parent, a quantity, and so on. The ATT (Attribute retrieval) category, for instance, draws its pool from [entity, attribute, value] triples, while the CNT (Counting) category draws from [class, number of occurrences] pairs. This pool is shuffled with a random number seeded from the run seed and the IFC file name. For a fixed IFC file, implementation version, and random seed, this procedure produces the same candidate ordering and option ordering. The generator then walks the shuffled pool up to a per-category cap and, for each fact, fills in a stem template, computes a set of distractors, and shuffles the option order so the correct answer is always at a random position.

\textit{(ii) Stem templates.} Each candidate item begins with a stem chosen from a fixed pool of templates defined for its category; the pools for the PRP (Property retrieval) and SPA (Spatial structure) categories are given below as an example. Placeholders such as \{prop\}, \{pset\}, \{entity\}, and \{file\} are filled with data read from the pool of facts created earlier. The template pools can be extended with additional wordings or structures as needed. \revision{The templates provide linguistic variation within each subtype rather than distinct retrieval operations. The operational coverage of the current benchmark is represented by the 15 categories and 36 subtypes, while items within a subtype intentionally instantiate the same retrieval operation for different entities, values, relationships, and source IFC models.}

\begin{itemize}
\footnotesize
    \item PRP (Property retrieval)
        \begin{itemize}[label=*]
            \item What is the value of the property \{prop\} in property set \{pset\} of {entity}?
            \item In the property set \{pset\} attached to {entity}, what value does \{prop\} hold?
            \item For \{entity\}, read \{prop\} from property set \{pset\}. What is its value?
        \end{itemize}
    \item SPA (Spatial structure)
    \begin{itemize}[label=,leftmargin=*]
        \item Element-level
        \begin{itemize}
            \item Which building storey directly contains \{entity\}?
            \item In the spatial structure of \{file\}, \{entity\} is placed on which storey?
            \item Which storey does \{entity\} belong to?
        \end{itemize}

        \item Space-level
        \begin{itemize}
            \item Which building storey is the space \{entity\} part of?
            \item The space \{entity\} belongs to which storey of \{file\}?
        \end{itemize}

        \item Building-level
        \begin{itemize}
            \item Which building aggregates the storey \{entity\}?
            \item In the spatial hierarchy of \{file\}, \{entity\} is a part of which building?
        \end{itemize}
    \end{itemize}
    
\end{itemize}

\textit{(iii) Distractors.} IFC-MCQ constructs distractors that reflect plausible IFC retrieval and reasoning errors. Wherever possible, distractors are derived from the source IFC file. For attribute, property, quantity, material, and spatial questions, they are typically values taken from sibling entities, representing the error of retrieving the correct field from the wrong entity. Relationship questions use incorrect hosts, incorrect fillers, objects of the wrong type, or incomplete material sets drawn from elsewhere in the same building model. Quantity questions may include values expressed in an incorrect unit, counting questions use nearby integers, and aggregation questions include values resulting from omitted instances or from applying the wrong aggregation operation, such as a sum instead of a mean. When the file does not provide enough distinct alternatives, the option-set is completed using a small, fixed set of transformations of the correct value. Existence / absence (EXS) items include ``Not specified in the file'' where applicable. A candidate is skipped when a valid set of distinct options cannot be formed.

\textit{(iv) Metadata.} Each candidate also carries a metadata record that captures the fact the item was built from, such as the identifier of the entity involved, the property or relationship queried, and the expected value or related entity. For instance, the ATT (Attribute retrieval) category, which builds its candidate pool from [entity, attribute, value] triples, stores a metadata record of [GUID of the entity, attribute, expected value]. Such a record lets an item be checked without parsing its natural language form, and it serves as the interface between the generator and the verifier.

\subsubsection{Verifiers and Structural Checks}
\label{sec:method-verify}

A generated item remains a candidate until its answer has been separately verified, so that generation errors do not propagate into the benchmark. \revision{IFC-MCQ verifies every candidate through a separate code path that sees only the item metadata and never the stored answer key(s). The verifier re-opens the source IFC file, resolves the referenced entities (typically by GlobalId), and re-derives the answer.} Depending on the category, it uses \textit{IfcOpenShell}\footnote{https://ifcopenshell.org/} utility functions, a fresh inverse relationship traversal, a recount with SI unit reconversion, or a coarser mesh deflection than the generator used. An item is discarded when the two answers disagree. \revision{Because generation and verification share the same \textit{IfcOpenShell} build, this check does not guard against a library bug, but it does catch generator-side counting, unit, and lookup errors.}

Items that pass verification go through a further round of structural checks. These confirm that each item has a unique identifier, that its options are distinct, and that the correct answer appears among them; an item is discarded if any distractor matches the correct answer within numeric tolerance. Each question type also has to match its expected shape, for instance a boolean question must have exactly two options. Only items that clear both verification and these checks are kept in the benchmark.

\section{Implementation and Evaluation Setup}
\label{sec:method-eval}

The IFC-MCQ pipeline is implemented as a Python software library built on top of \textit{IfcOpenShell} (v0.8.5). The implementation is designed to be model-agnostic and was tested on the ten parseable IFC files described below. This section provides details on the implementation and evaluation setup.

\subsection{IFC Models}
\label{sec:benchmark}

\begin{figure}
    \centering
    \begin{subfigure}[b]{0.24\linewidth}
        \centering
        \includegraphics[width=\linewidth]{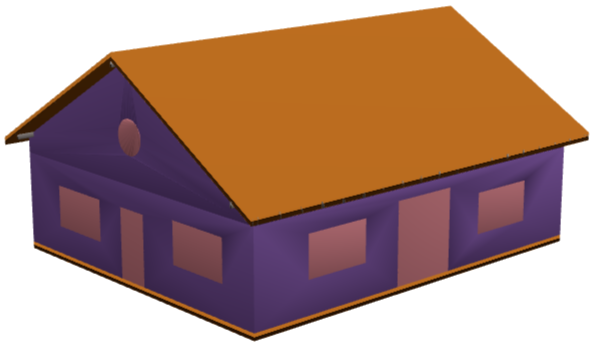}
        \caption{FZK Haus}
        \label{fig:fzk-haus}
    \end{subfigure}
    \hfill
    \begin{subfigure}[b]{0.24\linewidth}
        \centering
        \includegraphics[width=\linewidth]{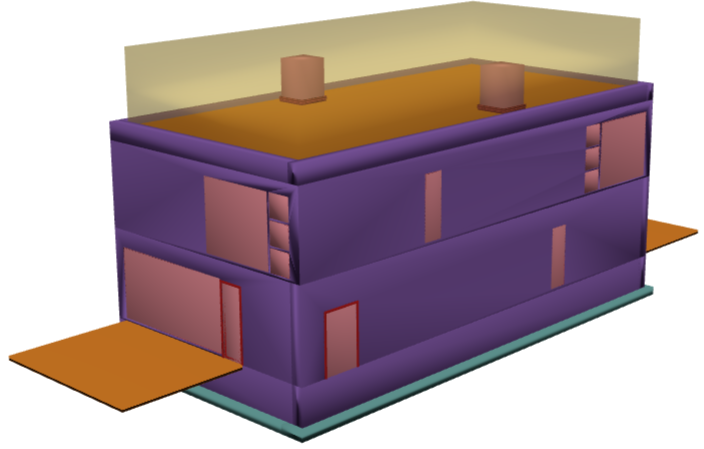}
        \caption{Duplex}
        \label{fig:duplex}
    \end{subfigure}
   \hfill
    \begin{subfigure}[b]{0.24\linewidth}
        \centering
        \includegraphics[width=0.8\linewidth]{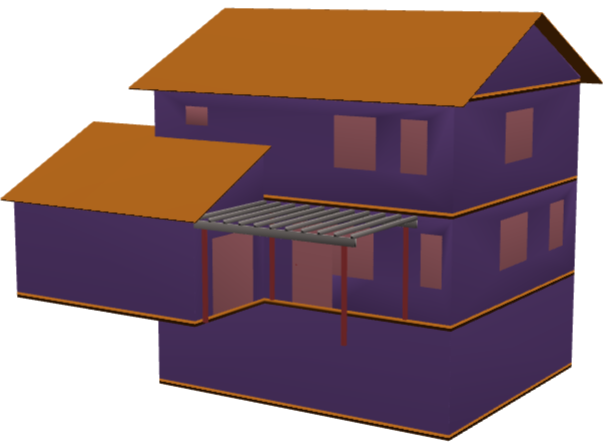}
        \caption{NiedriHaus}
        \label{fig:niedriHaus}
    \end{subfigure}
    \hfill
    \begin{subfigure}[b]{0.24\linewidth}
        \centering
        \includegraphics[width=0.6\linewidth]{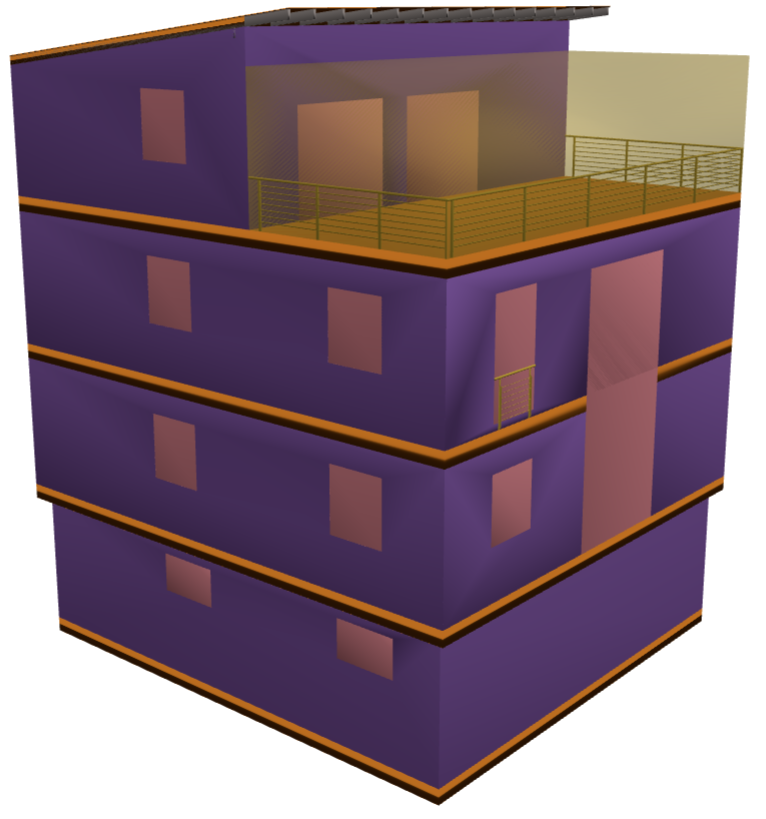}
        \caption{IAI Demo}
        \label{fig:iai-demo}
    \end{subfigure}

    \vspace{0.5em}

    \begin{subfigure}[b]{0.48\linewidth}
        \centering
        \includegraphics[width=0.8\linewidth]{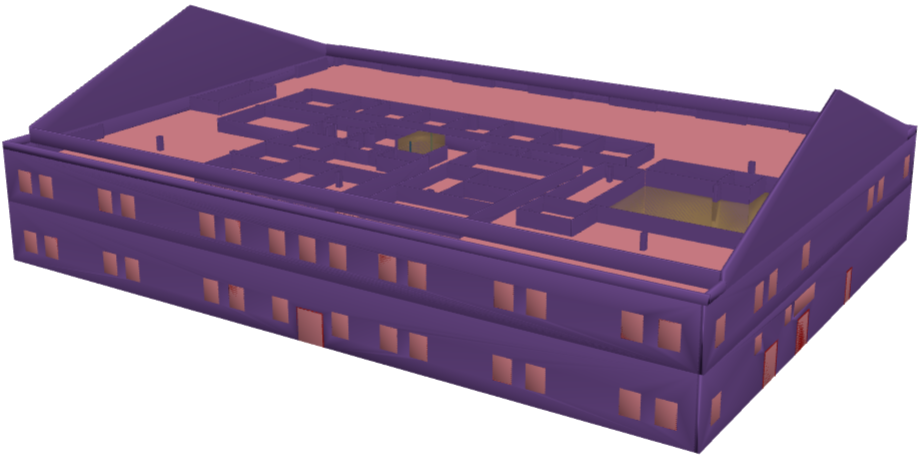}
        \caption{Office}
        \label{fig:office}
    \end{subfigure}
    \hfill
    \begin{subfigure}[b]{0.48\linewidth}
        \centering
        \includegraphics[width=0.8\linewidth]{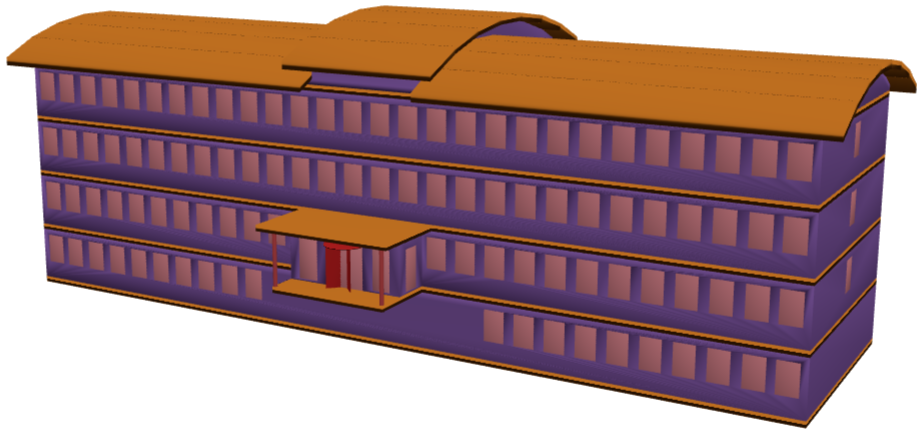}
        \caption{Institute}
        \label{fig:institute}
    \end{subfigure}

    \vspace{0.5em}

    \begin{subfigure}[b]{0.48\linewidth}
        \centering
        \includegraphics[width=\linewidth]{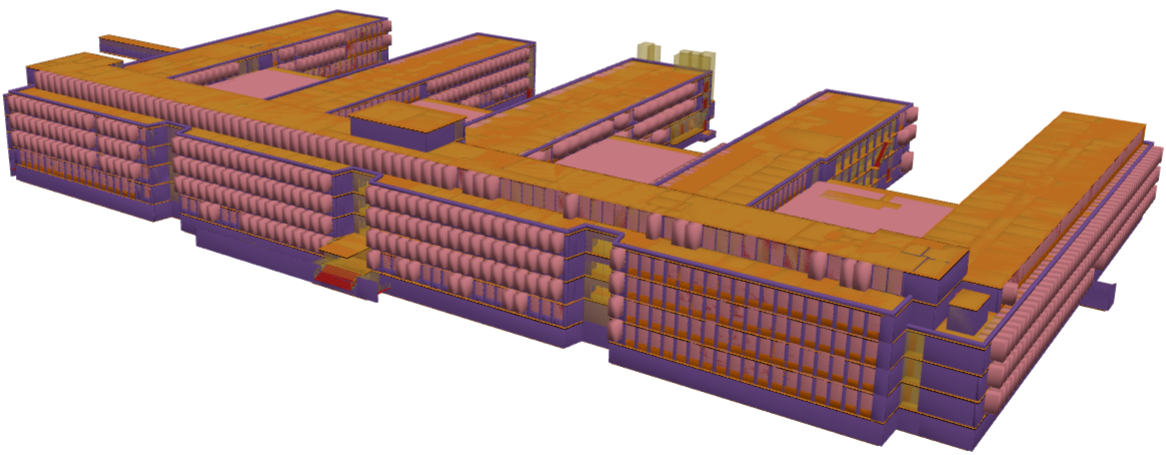}
        \caption{TNW}
        \label{fig:tnw}
    \end{subfigure}
    \hfill
    \begin{subfigure}[b]{0.48\linewidth}
        \centering
        \includegraphics[width=1.2\linewidth]{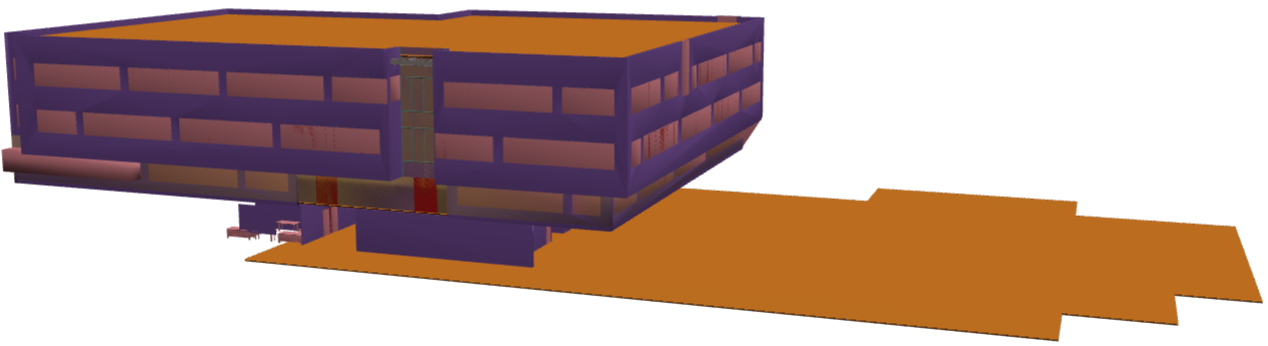}
        \caption{Trapelo}
        \label{fig:trapelo}
    \end{subfigure}

    \vspace{0.5em}

    \begin{subfigure}[b]{0.48\linewidth}
        \centering
        \includegraphics[width=\linewidth]{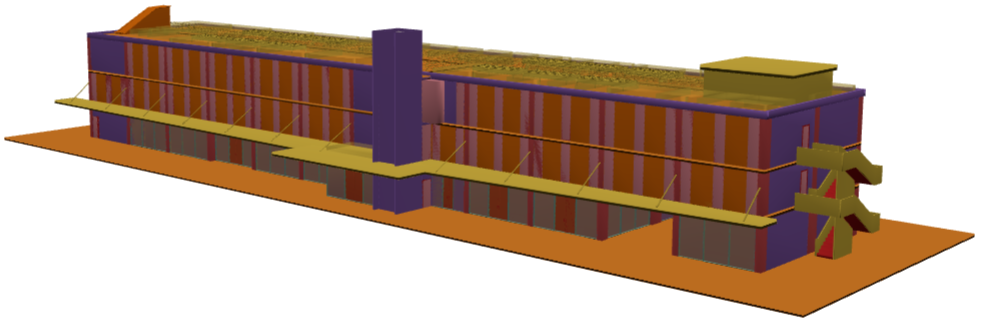}
        \caption{UNSW Block House}
        \label{fig:unsw-block-house}
    \end{subfigure}
    \hfill
    \begin{subfigure}[b]{0.48\linewidth}
        \centering
        \includegraphics[width=1.2\linewidth]{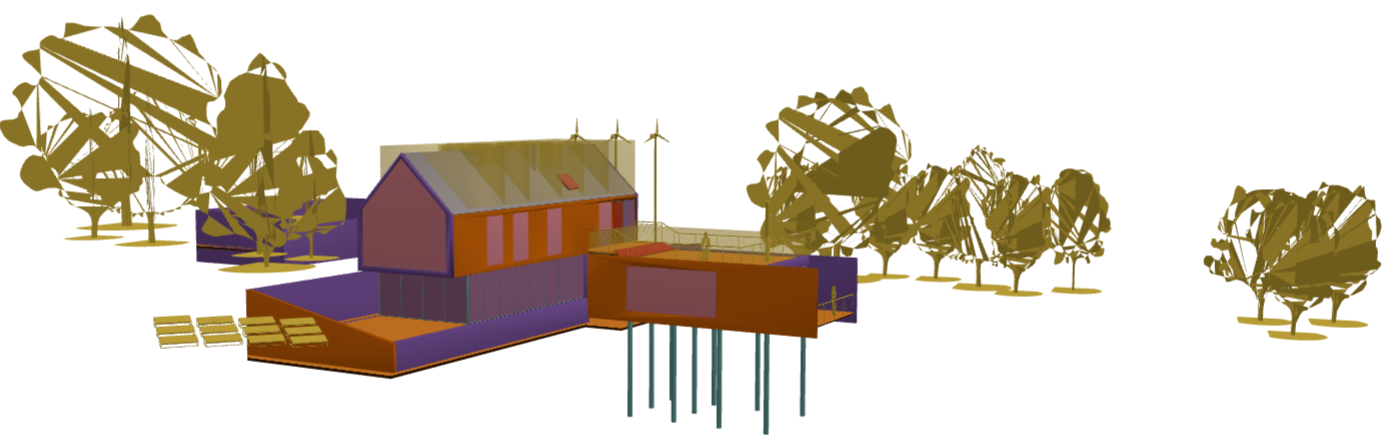}
        \caption{Sample House}
        \label{fig:img08}
    \end{subfigure}

    \caption{Rendered views of the ten IFC models used in this study. The views are illustrative and are not shown at a common scale. All models are the property of their respective authors; none were created in this work.}
    \label{fig:all-models}
\end{figure}

\begin{table}
\centering
\footnotesize
\caption{Summary of IFC building models and their contained elements. The reported statistics are based on counts of \textit{IfcBuildingStorey}, \textit{IfcElement}, \textit{IfcSpace}, \textit{IfcWall}, \textit{IfcSlab}, \textit{IfcWindow}, and \textit{IfcDoor} entities.}
\label{tab:ifc-statistics}
\begin{tabular}{@{}lrrrrrrr@{}}
\toprule
\textbf{IFC Model} & \textbf{Storeys} & \textbf{Elements} & \textbf{Spaces} & \textbf{Walls} & \textbf{Slabs} & \textbf{Windows} & \textbf{Doors} \\ \midrule
FZK Haus        & 2  & 131   & 7    & 13    & 4  & 11   & 5    \\
Duplex          & 4  & 268   & 21   & 57    & 21 & 24   & 14   \\ \midrule
NiedriHaus      & 4  & 146   & 19   & 45    & 8  & 16   & 16   \\
IAI Demo        & 4  & 189   & 21   & 38    & 6  & 30   & 22   \\
Office          & 3  & 986   & 99   & 487   & 2  & 69   & 102  \\
Institute       & 5  & 989   & 82   & 119   & 26 & 206  & 77   \\
TNW             & 11 & 23,053 & 1,103 & 10,310 & 99 & 1,725 & 1,271 \\
Trapelo         & 5  & 3,583  & 139  & 522   & 60 & 68   & 122  \\
UNSW Block House & 6 & 1,494 & 105 & 292 & 6 & 128 & 105 \\
Sample House & 6 & 502 & 14 & 48 & 13 & 17 & 21 \\ 
\bottomrule
\end{tabular}
\end{table}

\revision{We apply the IFC-MCQ pipeline to ten IFC models of varying sizes shown in Figure \ref{fig:all-models}; the summary of the IFC models and their contained elements is provided in Table \ref{tab:ifc-statistics}. Benchmark generation uses a random seed of 10 and a per-category cap of 40 verified items per IFC model. The cap prevents categories with large candidate pools from disproportionately dominating the generated benchmark while retaining sufficient within-category variation. It is a balancing parameter rather than a theoretical upper bound on the number of items that IFC-MCQ can generate. We additionally use a reporting threshold of 10 verified items to identify categories with limited representation in a given IFC model.} For single-choice questions, the number of distractors ranges from two to four, whereas multi-choice questions require at least one distractor. The mesher deflection tolerance is set to 0.001 for generation and 0.02 for verification.

\subsection{Agentic System Evaluation}
\label{sec:agentic-system}

\begin{figure}
    \centering
    \includegraphics[width=\linewidth]{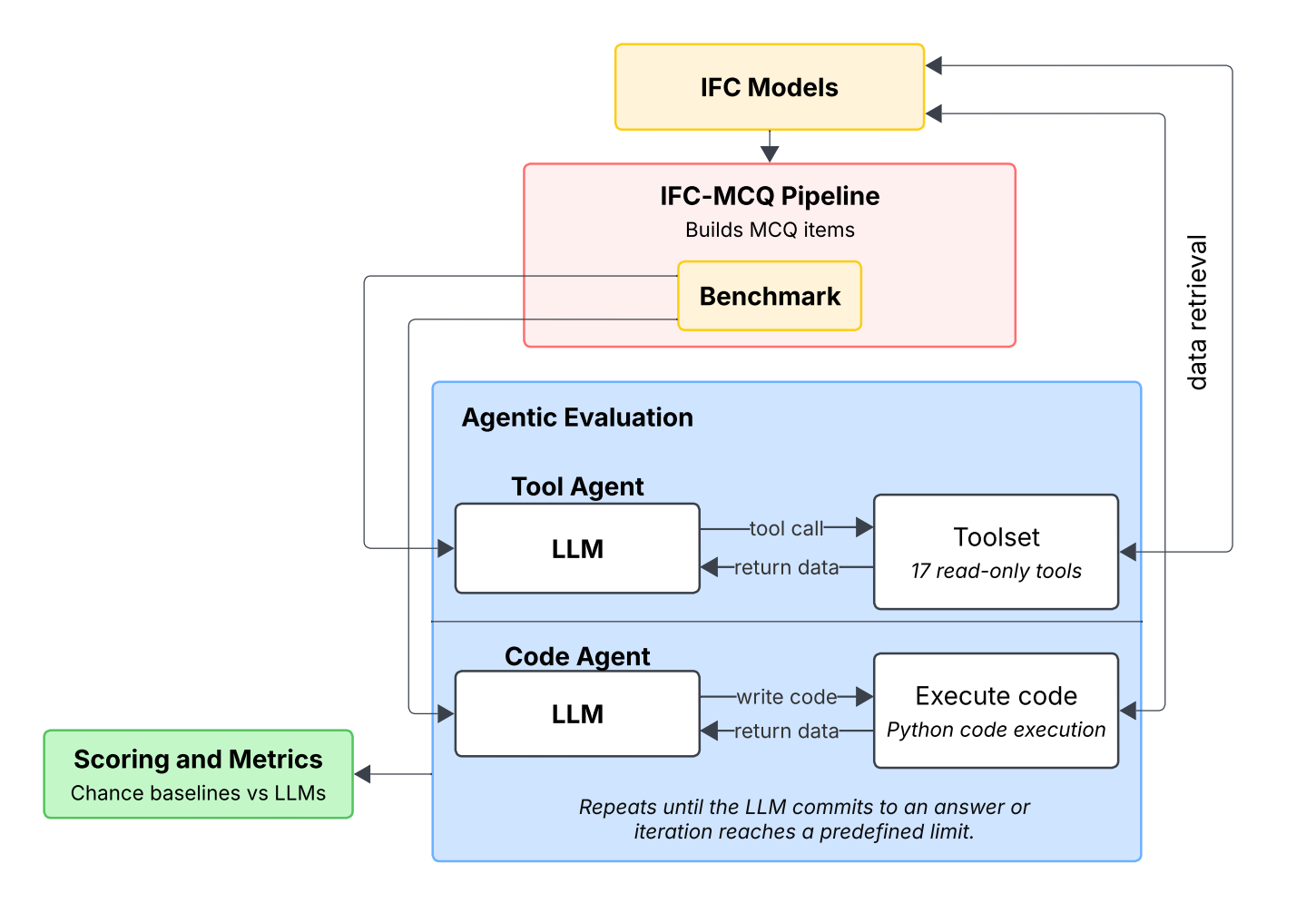}
    \caption{Overview of the agentic evaluation. The tool agent retrieves information through a fixed set of 17 read-only tools, while the code agent writes and executes IfcOpenShell code against the IFC model.}
    \label{fig:agentic-evaluation}
\end{figure}

\revision{We evaluate multiple LLMs using two agentic configurations relevant to existing IFC retrieval approaches: a tool agent and a code agent (Figure~\ref{fig:agentic-evaluation}). Throughout the evaluation, the prompts, protocol, and scoring remain identical across IFC models, so that any difference in performance reflects only the underlying LLM and the interaction mode of the agent. Before answering a query, each agent must inspect the building model in question.} This design mirrors how an LLM-based agent would operate on a real project: it has no prior knowledge of the model and must retrieve the relevant facts by traversing the entities and relationships encoded in the IFC file, as required by the question at hand. Both agents operate through repeated retrieval and reasoning for a maximum of eight interaction steps. At each step, the LLM either performs one permitted action or returns a final answer. If no parseable answer is produced within eight steps, the LLM receives one additional answer-only prompt requiring a response in the prescribed format.

\begin{figure}
    \centering
    \includegraphics[width=0.9\linewidth]{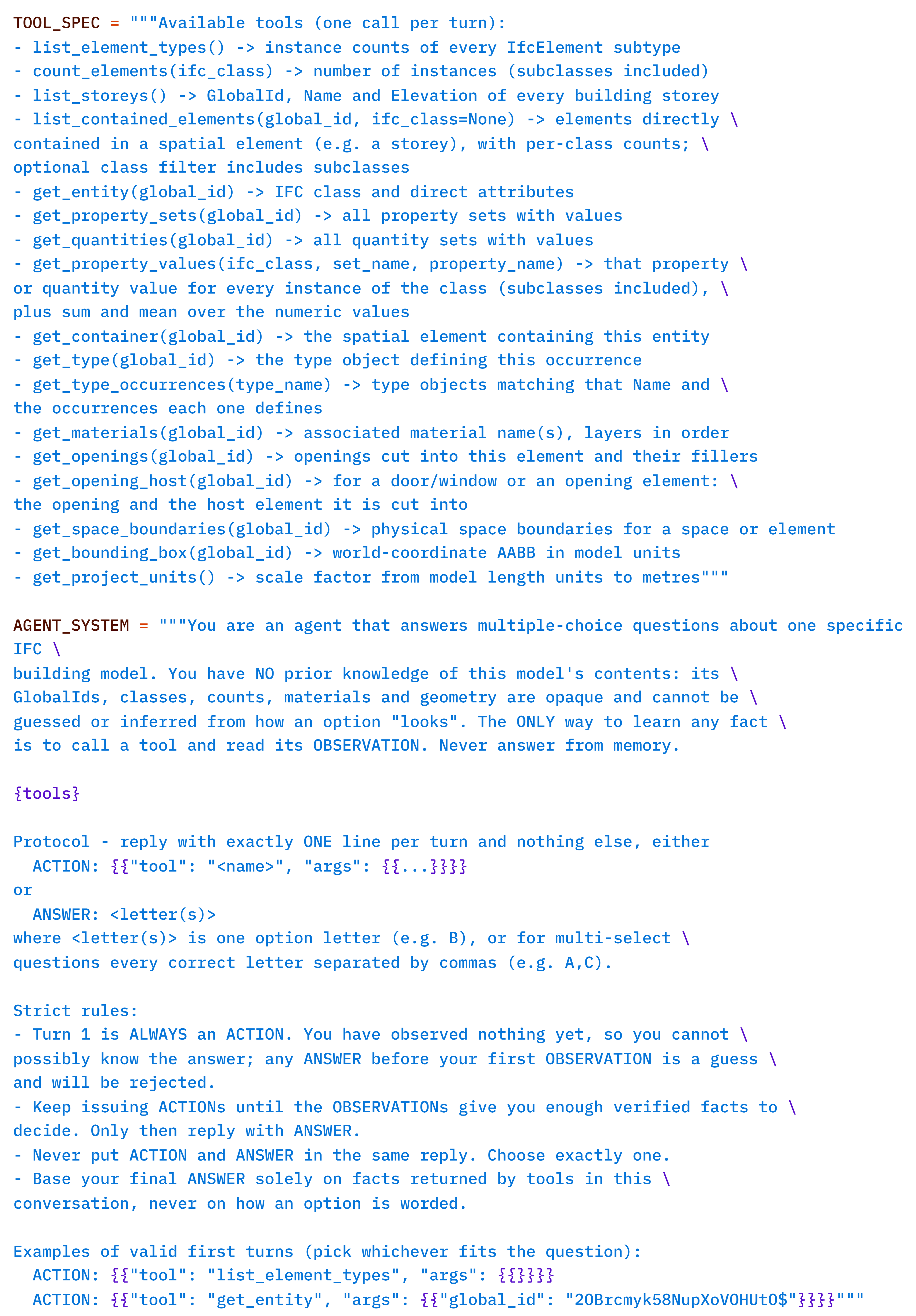}
    \caption{System prompt template used by the tool agent.}
    \label{fig:system-prompt}
\end{figure}

\paragraph{Tool agent}
\revision{The tool agent can access the IFC file only through a set of 17 read-only tools, listed in Figure \ref{fig:system-prompt}. The tool set was designed based on the question taxonomy and provides at least one retrieval route for every category in the current IFC-MCQ scope. It is therefore operationally complete with respect to the present benchmark. Some tools expose stored IFC facts, while others perform category-relevant aggregation or geometric operations before returning an observation. The tool agent evaluation consequently measures whether an LLM can select an appropriate operation, supply the correct arguments, interpret the resulting observation, and construct the required answer set. It does not require the LLM to synthesize the underlying IFC traversal or computation from first principles. To bound context growth during agent interaction, each tool observation is limited to 6,000 characters. When an observation exceeds this limit, its content is truncated before being supplied to the LLM. The tools are implemented as Python functions. At each step, the LLM either calls a tool and receives the returned data or provides its final answer. Figure \ref{fig:system-prompt} presents the system prompt template, including the tool specifications and instructions.}

\begin{figure}
    \centering
    \includegraphics[width=0.9\linewidth]{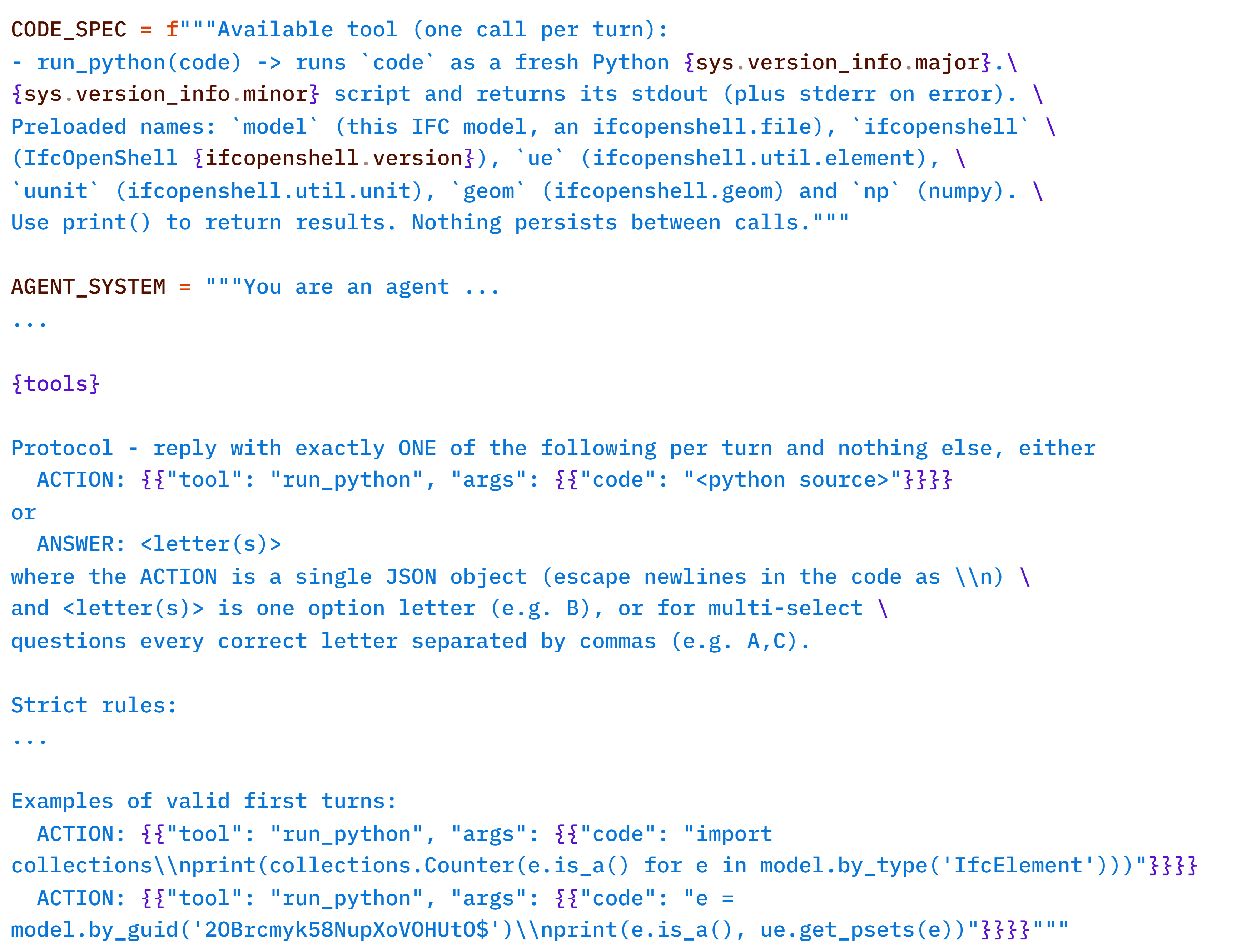}
    \caption{System prompt template used by the code agent.}
    \label{fig:code-agent-system-prompt}
\end{figure}

\paragraph{Code agent}
\revision{In contrast to the fixed tool set, the code agent has no predefined retrieval functions. This condition removes the task-specific abstraction provided by the tool agent. Instead, the LLM must write Python code using the \textit{IfcOpenShell} library to query the IFC model directly. It must identify the relevant IFC structures, select appropriate \textit{IfcOpenShell} operations, compose executable code, recover from runtime errors, and interpret the output. The generated code is executed in an isolated container against the IFC models, and the resulting output or error message is returned to the LLM. This allows the agent to compose its own queries, for example by combining several traversals or filters in a single step, and to recover from errors by revising its code. As with tool observations, code-execution output was limited to 6,000 characters and was truncated when this limit was exceeded. The comparison therefore examines how performance changes when the retrieval procedure itself must be synthesized rather than selected from a taxonomy-complete tool set. Figure \ref{fig:code-agent-system-prompt} presents the system prompt template used by the code agent.}

\paragraph{LLMs Under Evaluation}
Table~\ref{tab:llms} lists the LLMs evaluated in this work. The selection was intended to provide broad coverage of currently available instruction-tuned LLMs \citep{zhang2026instruction}, with emphasis on: (i) representation across multiple major LLM developers, (ii) variation in LLM size, (iii) both dense and mixture-of-experts (MoE) architectures \citep{cai2025survey}, and (iv) both proprietary and open-weight deployment paradigms. Within these constraints, we selected LLMs that were accessible through an inference API or available as downloadable open-weight releases as of July 2026. The resulting set includes models from OpenAI \citep{agarwal2025gpt,openai2026gpt54mininano,openai2026gpt56}, DeepSeek \citep{xu2026deepseek}, Google DeepMind \citep{team2025gemma,team2026gemma}, Meta \citep{grattafiori2024llama,meta2025llama}, Alibaba Cloud \citep{qwenteam2026qwen3627b,qwen37}, Moonshot AI \citep{team2026kimi}, MiniMax \citep{lai2026minimax}, NVIDIA \citep{blakeman2025nemotron}, and Z.ai \citep{zeng2026glm}. The open-weight models range from compact (8--30.7 billion parameters) to large-scale (117 billion--2.8 trillion parameters) configurations.

\begin{table}[t!]
    \centering
    \footnotesize
    \caption{LLMs evaluated in this work. $^{\dagger}$Parameter count has not been publicly disclosed.}
    \label{tab:llms}
    \begin{tabular}{l|c}
    \toprule
    \textbf{LLM} & \textbf{Size (params)}\\
    \midrule
    \multicolumn{2}{l}{\textit{Proprietary frontier models}}\\
    \midrule
    qwen3.7-max \citep{qwen37}  & n/a$^{\dagger}$\\
    gpt-5.6-luna \citep{openai2026gpt56}  & n/a$^{\dagger}$\\
    \hdashline
    \textit{compact} & \\

    gpt-5.4-mini \citep{openai2026gpt54mininano}  & n/a$^{\dagger}$\\
    gpt-5.4-nano \citep{openai2026gpt54mininano}  & n/a$^{\dagger}$\\
        \hdashline
    \midrule
    \multicolumn{2}{l}{\textit{Large open-weight models}}\\
    \midrule
    kimi-k3 \citep{team2026kimi}  & 2.8T (104B active)\\
    deepseek-v4-pro \citep{xu2026deepseek}  & 1.6T (49B active)\\
    glm-5.2 \citep{zeng2026glm}  & 744B (40B active)\\
    minimax-m3 \citep{lai2026minimax}  & 428B (23B active)\\
    llama-4-maverick \citep{meta2025llama} & 400B (17B active)\\
    deepseek-v4-flash \citep{xu2026deepseek}  & 284B (13B active)\\
    gpt-oss-120b \citep{agarwal2025gpt}  & 117B (5.1B active)\\
    \midrule
    \multicolumn{2}{l}{\textit{Compact open-weight models}}\\
    \midrule
    gemma-4-31b-it \citep{team2026gemma}  & 30.7B\\
    nemotron-3-nano-30b-a3b \citep{blakeman2025nemotron}  & 30B (3.5B active)\\
    gemma-3-27b-it \citep{team2025gemma}   & 27B\\
    qwen3.6-27b \citep{qwenteam2026qwen3627b}  & 27B\\
    gpt-oss-20b \citep{agarwal2025gpt}  & 21B (3.6B active)\\
    llama-3.1-8b-instruct \citep{grattafiori2024llama} &   8B\\
    \bottomrule
    \end{tabular}
\end{table}

\subsubsection{Scoring and metrics}
An answer is scored as correct only if the selected option(s) exactly match the answer key(s). This means that the LLM under evaluation must produce an exact match, following the format \texttt{ANSWER: <letter(s)>}, where \texttt{<letter(s)>} is a single letter (e.g., B) for questions with one correct answer, or a comma-separated list of letters (e.g., A,C) for multi-choice questions. The order of the selected options is irrelevant, as we compare sorted letter sequences during evaluation. We report accuracy overall and by group, category, subtype, and question type. To put model performance in context, we also evaluate two \textit{chance baselines}: one that always selects the first option (\textit{dummy}) and one that selects options at random (\textit{random}). Together, these baselines give an estimate of the accuracy achievable by chance alone.

\paragraph{Inference settings and failure handling} Inference was conducted using the \textit{OpenAI API library}\footnote{\url{https://pypi.org/project/openai/}}, which is natively supported by several inference service providers. \revision{OpenAI and DeepSeek models were accessed through their native APIs, whereas the remaining models, including GPT-OSS, were served through Novita AI. For all models that supported the parameter, the temperature was set to 0. All other parameters were left at their provider defaults, including the maximum token limit and reasoning efforts. Unsuccessful actions, including calls with an unknown tool name, invalid arguments, or an \textit{IfcOpenShell} exception, were returned to the LLM as observations. The agent could therefore inspect the error and retry with a revised call on a subsequent step; unsuccessful calls were not manually corrected or removed from the interaction.}

\subsubsection{Evaluation Protocol}
\label{sec:evaluation-protocol}
\revision{The evaluation proceeds in three stages designed to examine LLM performance, variation across source IFC models, and the effect of the retrieval interface.} 

\revision{\textit{Stage 1} evaluates all 17 LLMs (listed in Table \ref{tab:llms}) with the tool agent on the benchmarks generated from FZK Haus and Duplex. Both models are relatively small and provide broad coverage of the IFC-MCQ categories. The tool agent also has a bounded action space, since each step selects one of 17 predefined tools. \textit{Stage 1} therefore provides a controlled baseline for comparing LLMs, identifying category-level failure modes, and establishing performance relative to the chance baselines before introducing more heterogeneous IFC models and a less constrained retrieval interface.}

\revision{\textit{Stages 2} and \textit{3} use a subset of five LLMs because inference cost increases with the number of evaluated models, IFC files, and agent configurations. The subset was selected after Stage 1 to cover the main design factors and the observed range of Stage 1 performance. It includes one proprietary model (\textit{gpt-5.6-luna}), two large open-weight mixture-of-experts models (\textit{deepseek-v4-flash} and \textit{gpt-oss-120b}), and two compact dense models (\textit{gemma-3-27b-it} and \textit{qwen3.6-27b}), with Stage 1 accuracies ranging from 87.4\% to 100\%. \textit{llama-3.1-8b-instruct}, which did not reliably execute the agent loop in Stage 1, was not carried forward. Stage 2 retains the tool agent and extends the evaluation to four additional IFC models (Institute, Office, Trapelo, and TNW), giving six models ranging from 131 to 23k building elements, selected to broaden the range of IFC model sizes. Stage 3 evaluates the code agent on the same six IFC models with the same five LLMs, allowing the effect of the retrieval interface to be examined while holding the LLMs and evaluation models fixed. The scoring procedure, eight-interaction-step limit, and final answer-only fallback remain unchanged.}

\revision{For Stages 2 and 3, up to 10 verified items are selected from each represented category for each IFC model. This fixed per-category quota prevents categories with larger generated pools from contributing disproportionately to the evaluation and limits the inference cost associated with evaluating multiple LLMs and agent configurations. FZK Haus, Duplex, Institute, Office, and Trapelo each contribute 150 items, while TNW contributes 140 items because it contains no Space boundaries (BND) items, giving 890 items in total. The same 890 items, including the same question text, options, and option ordering, are used for both the tool and code agents.}

\paragraph{No-IFC-access control}
\revision{To estimate how much performance could be achieved without retrieving information from the source IFC models, we evaluated \textit{gpt-5.6-luna}, \textit{qwen3.6-27b}, \textit{gemma-3-27b-it}, and \textit{llama-3.1-8b-instruct} on the benchmark used in Stages 2 and 3. These models were selected to span the range of performance observed across the agentic evaluations, including the lowest-performing model in Stage 1. Each LLM received only the question stem and answer options, without access to the IFC files, retrieval tools, code execution, or retrieved observations. Responses used the same format and exact option-set grading procedure as in the agentic evaluation.}

\subsection{Human audit of generated items}
\label{sec:human-audit-method}

\revision{Further, to complement the automated generation and verification procedures, we conducted a manual audit of a stratified sample from the benchmark used in Stages 2 and 3. One item was selected uniformly at random from each represented combination of IFC model and question category using a fixed random seed of 10. FZK Haus, Duplex, Institute, Office, and Trapelo each contributed 15 items, whereas TNW contributed 14 because it does not contain Space boundaries (BND) items.}

\revision{The audit was conducted by one human evaluator in two steps. First, the evaluator was provided with the question text and source IFC model, but not the answer options or benchmark answer key. The evaluator checked whether the question was grammatically well formed, including whether all template placeholders had been instantiated correctly, and independently derived the answer by inspecting the IFC model. Second, the answer options were revealed. The evaluator then checked whether the independently derived answer was present and uniquely represented among the options, with no duplicated answer option. For multi-choice items, every member of the independently derived answer set had to be represented exactly once, and no additional option could also be supported by the source IFC model. The human-derived answer was compared with the benchmark answer key only after these checks had been completed.}

\section{Results}
\label{sec:results}
\revision{This section first reports benchmark-generation and verification results across the ten IFC models, followed by the human-audit results (Section~\ref{sec:generation-results}). It then presents the three-stage agentic evaluation. Stage 1 evaluates the tool agent on FZK Haus and Duplex (Section~\ref{tool-agent-fzk-duplex}); Stage 2 examines tool agent performance across these two models and four additional IFC models (Section~\ref{tool-agent-additional-models}); and Stage 3 compares the tool and code agents using the same six models, LLMs, and benchmark items (Section~\ref{code-agent-all-models}). Finally, Section~\ref{sec:no-ifc-control} reports the no-IFC-access control.}

\subsection{Benchmark Generation, Verification, and Human Audit}
\label{sec:generation-results}

This section reports the generation and verification results for the ten IFC models discussed in Section \ref{sec:benchmark}, and the stratified human audit.

\paragraph{FZK Haus and Duplex} These two reference IFC models are relatively small and structurally simple. We ran the pipeline on them first, before extending it to the remaining eight models. In total, 1,177 verified items were generated; 596 from the FZK Haus and 581 from the Duplex, and every item passed verification. Most categories reached the cap of 40 items per model (80 in total). The exceptions were Quantity retrieval (QTY, 61 items) and Aggregation (AGG, 76 items), which are limited by the number of admissible quantity facts in each model. Of the 36 subtypes in the taxonomy (Table~\ref{tab:subtypes}), 35 are represented in the benchmark. The SPA \textit{storey's building} subtype yielded no items for either model, because each file contains only a single building and no alternative buildings are available as distractors. By question type, there are 896 single-choice, 176 boolean, and 105 multi-choice items. Representative items are shown in Table~\ref{tab:sample-items}.

\begin{table}
\centering
\caption{Representative items from FZK Haus and Duplex. Correct answers are marked with an asterisk.}
\label{tab:sample-items}
\scriptsize
\setlength{\tabcolsep}{3.5pt}
\begin{tabular}{llp{7cm}p{6cm}}
\toprule
\textbf{Cat.} & \textbf{Type} & \textbf{Question} & \textbf{Options} \\
\midrule
CNT & S
  & Count the \texttt{IfcSlab} instances (subclasses included) in \texttt{FZK-Haus.ifc}.
  & (A)~15 \quad (B)~4$^*$ \quad (C)~1 \quad (D)~88 \quad (E)~7 \\
\addlinespace
CLS & S
  & What is the IFC class of the entity
    (GlobalId \texttt{2OrWItJ6zAwBNp0OUxK\$CR}) in \texttt{Duplex.ifc}?
  & (A)~\texttt{IfcRoof} \quad (B)~\texttt{IfcSlab} \newline (C)~\texttt{IfcFooting} \quad (D)~\texttt{IfcWall} \newline (E)~\texttt{IfcBeam}$^*$ \\
\addlinespace
SPA & S
  & Which storey does \texttt{IfcBeam}
    (GlobalId \texttt{3sG\$iUZAX499eUK87UfZk\_}, Name ``Sparren-17'') belong to?
  & (A)~\texttt{IfcSite} ``Gelaende'', \newline(B)~\texttt{IfcBuildingStorey} ``Dachgeschoss''$^*$, \newline(C)~\texttt{IfcBuilding} ``FZK-Haus'', \newline(D)~\texttt{IfcBuildingStorey} ``Erdgeschoss''\\
\addlinespace
ATT & S
  & What is the value of the \texttt{OverallWidth} attribute of \texttt{IfcDoor}
    (GlobalId \texttt{2qiPPF3FrF8OIqfrKiSUqm}, Name ``Innentuer-3'')?
  & (A)~1.01 \quad (B)~2.01 \quad (C)~0.885$^*$ \newline (D)~2 \quad (E)~1\\
\addlinespace
QTY & S
  & Read the quantity ``Length'' from quantity set ``BaseQuantities'' of
    \texttt{IfcWallStandardCase}
    (GlobalId \texttt{3Ttjr\$59XEWfWN1WUHjelZ}, Name ``Wand-Ext-OG-4'').
    What is its value in metres?
  & (A)~3.8\,m \quad (B)~3\,m \quad (C)~4.08\,m \newline(D)~10\,m \quad (E)~12\,m$^*$\\
\addlinespace
PRP & S
  & For \texttt{IfcBeam} (GlobalId \texttt{31Quw2HAvFqv\$9\_JgCuemp},
    Name ``Sparren-6''), read ``Slope'' from property set
    ``Pset\_BeamCommon''. What is its value?
  & (A)~0 \quad (B)~34.5 \quad (C)~42 \newline(D)~25.5 \quad (E)~30$^*$ \\
\addlinespace
BLN & B
  & For \texttt{IfcBeam} (GlobalId \texttt{0DTRkEz0r6gvzG6VdDr24n},
    Name ``Sparren-36''), is ``LoadBearing'' in property set
    ``Pset\_BeamCommon'' equal to True?
  & (A)~Yes$^*$ \quad (B)~No \\
\addlinespace
GEO & B
  & Is the point $(6.0, 2.25, 4.615)$ (world coordinates, metres) inside the
    axis-aligned bounding box of \texttt{IfcSlab}
    (GlobalId \texttt{2IxUUNUVPB6Ob\$eicCfP2N}, Name ``Dach-2'')?
  & (A)~True$^*$ \quad (B)~False \\
\addlinespace
EXS & B
  & Does \texttt{IfcBeam} (GlobalId \texttt{1nGDFBBHvBJ9NmlweQO0fX},
    Name ``Sparren-9'') have a value recorded for the property ``Slope''
    in property set ``Pset\_BeamCommon''?
  & (A)~Yes$^*$ \quad (B)~No \\
\addlinespace
MAT & M
  & Select every material that appears in the material (layer) set of
    \texttt{IfcCovering} (GlobalId \texttt{3bXiCStxP6Fgxdej\$yc55k},
    Name ``Compound Ceiling:Gypsum Board:187691'').
  & (A)~Metal - Stud Layer$^*$ \newline(B)~Masonry - Brick, \newline(C)~Wood - Flooring, \newline(D)~Plasterboard$^*$, \newline(E)~Masonry - Grout\\
\bottomrule
\end{tabular}
\end{table}

\paragraph{Additional IFC models}
\revision{We then applied IFC-MCQ to the remaining eight IFC models; results are summarized in Table \ref{tab:generation-results}. Across these models, 4,482 candidate items were generated, of which 4,467 passed verification and were retained. During candidate construction, 402 candidate-generation attempts were skipped because too few distinct answer options could be formed; all such skips occurred in Property retrieval (PRP) or Existence / absence (EXS). The resulting category coverage varied across the source models. NiedriHaus, for example, produced only one EXS item. EXS alternates between boolean and single-choice forms to maintain coverage across subtypes, and the single-choice form requires varied property values as distractors, which this model lacks. Likewise, Space boundaries (BND) produced no items for TNW or UNSW Block House because these files contain no \textit{IfcRelSpaceBoundary} relationships. Aggregation (AGG) category was also sparsely represented in UNSW Block House.}

\begin{table}
\centering
\footnotesize
\caption{Results from the IFC-MCQ pipeline across different IFC models. “Generated” denotes candidate items produced by the category generators, with the overall generation time shown in brackets. “Verified” denotes candidates retained after separate verification, with the overall verification time shown in brackets. “Skipped” denotes candidate-generation attempts for which a valid option-set could not be constructed. Categories with fewer than 40 items did not reach the generation cap; categories with fewer than 10 verified items fall below the reporting threshold.}
\label{tab:generation-results}
\begin{tabular}{p{2cm}llp{3cm}p{3cm}p{3cm}}
\toprule
\textbf{IFC Model}             & \textbf{Generated} & \textbf{Verified} & \textbf{Skipped during generation} & \textbf{10 to <40 items}                              & \textbf{<10 items} \\ \midrule
FZK Haus       & 596 (6.72s) & all (3.16s) & PRP (3)             & AGG (36)                     & -                         \\
Duplex         & 581 (4s) & all (1.42s) & PRP (181), EXS (68) & QTY (21)                     & -                         \\ \midrule
NiedriHaus & 495 (3.87s)        & all (1.88s)       & PRP (63), EXS (68)        & PRP (19), TYP (32), AGG (13), CMP (30) & EXS (1)     \\
IAI Demo       & 571 (3.81s) & all (2.42s) & EXS (12)            & AGG (19), CMP (32)           & -                         \\
Office         & 597 (9.02s) & all (2.83s) & PRP (98), EXS (28)  & CMP (37)                     & -                         \\
Institute & 589 (4.08s) & all (2.17s) & -                   & AGG (33), CMP (36)           & -                         \\
TNW            & 560 (1,146s) & 557 (30.66s) & PRP (20), EXS (11)  & EXS (39), CNT (39), AGG (39) & BND (0)                   \\
Trapelo        & 600 (26.61s) & 596 (5.26s) & PRP (8), EXS (7)    & CNT (36)                     & -                         \\
UNSW Block House           & 470 (250.34s) & all (80.24s) & PRP (25), EXS (5)   & CMP (24)                     & BND (0), QTY (0), AGG (6) \\
Sample House         & 600 (34.77s) & 592 (4.52s) & PRP (22), EXS (35)  & CNT (38), AGG (34)           & -                         \\ \bottomrule
\end{tabular}
\end{table}

\revision{Fifteen generated candidates across TNW, Trapelo, and Sample House were rejected during verification because the answer re-derived through the separate verification path did not match the generator result. The disagreements resulted from four implementation or modelling cases: (i) an existence check considered only the occurrence and missed a property stored on its type; (ii) a storey count omitted assembly components associated with the storey through their parent; (iii) an aggregation considered only the exact IFC class although the question included subclasses; and (iv) property sets sharing the same name were resolved differently. These cases demonstrate the value of applying the generation and verification pipeline to heterogeneous IFC models, since variations in modelling structure can expose edge cases that are not apparent in smaller reference models.}

\paragraph{Human audit results}

\revision{The stratified sample was selected as described in Section~\ref{sec:human-audit-method}. All 89 sampled items passed the three manual audit checks (Table~\ref{tab:human-audit}). Every question was judged to be grammatically well formed, and all template placeholders were instantiated correctly. For every item, the answer independently derived from the source IFC model matched the IFC-MCQ benchmark answer key. The corresponding answer or answer set was also present and uniquely represented among the options, with no duplicate option representing the correct answer. No sampled item required correction or removal.}

\begin{table}
\centering
\footnotesize
\caption{Results of the stratified human audit of generated benchmark items.}
\label{tab:human-audit}
\begin{tabular}{@{}lr@{}}
\toprule
\textbf{Audit criterion} & \textbf{Passed} \\
\midrule
Grammatically well formed and placeholders correctly instantiated
    & 89/89 (100\%) \\
Human-derived answer matched the benchmark answer key
    & 89/89 (100\%) \\
Answer uniquely represented among the options
    & 89/89 (100\%) \\
\bottomrule
\end{tabular}
\end{table}

\subsection{Stage 1: Tool agent on FZK Haus and Duplex}
\label{tool-agent-fzk-duplex}

This section presents the Stage 1 results at the overall, group, question-type, category, and subtype levels.

\subsubsection{Overall Performance}
\label{sec:results-overall}

\begin{figure}[t!]
    \centering
    \includegraphics[width=0.9\linewidth]{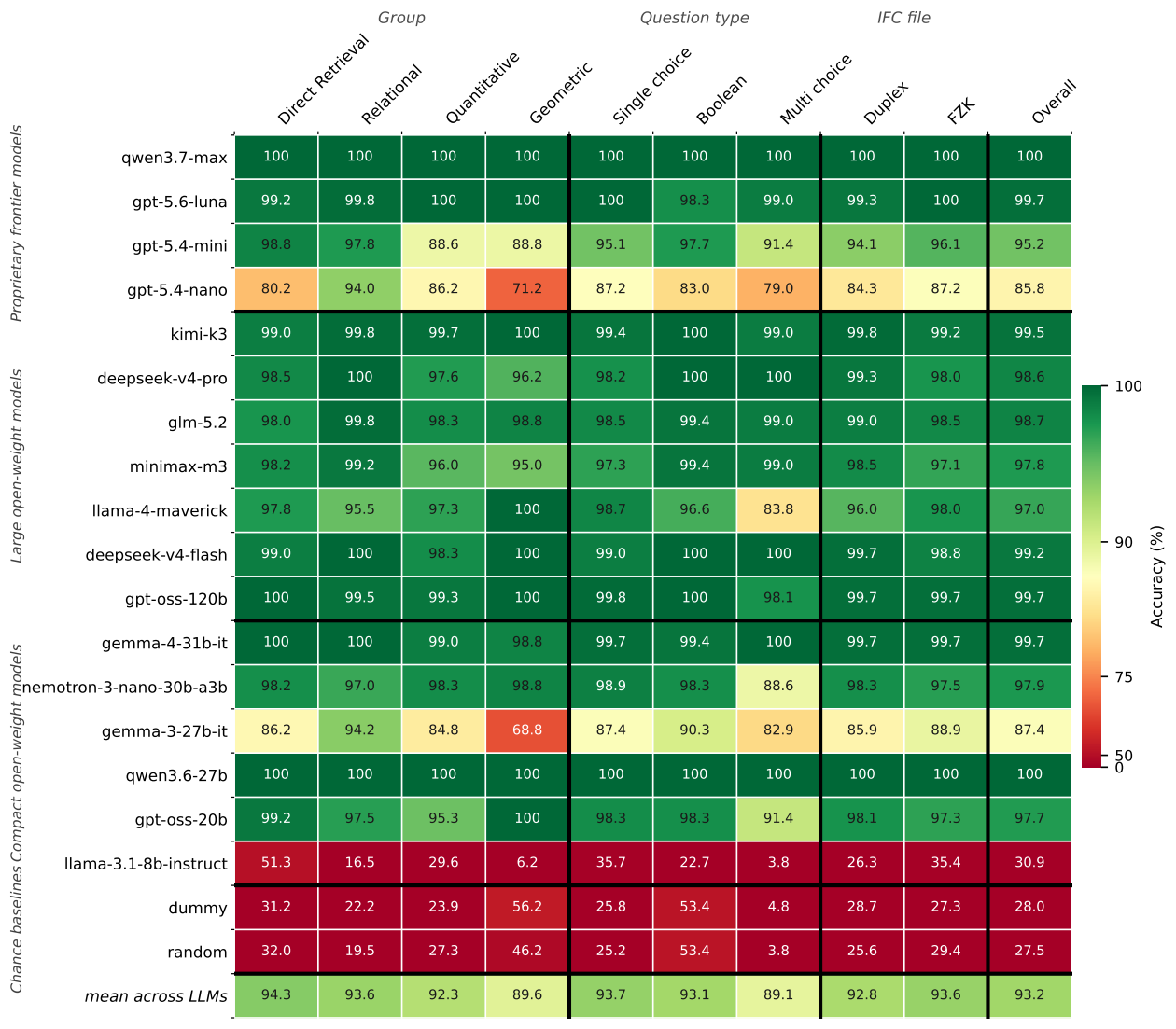}
    \caption{Performance of 17 LLMs using the tool agent on all 1,177 items generated from FZK Haus and Duplex, reported by question group, question type, and source IFC model.}
    \label{fig:acc-overall}
\end{figure}

All evaluated LLMs outperform the chance baselines. The dummy and random baselines reach 28.0\% and 27.5\%, respectively. These baselines measure fixed-position and random selection.  Among the 17 LLMs, 14 exceed 95\% overall accuracy and seven exceed 99\%. Two models, \textit{qwen3.7-max} and \textit{qwen3.6-27b}, answer every item correctly. Three others, \textit{gpt-5.6-luna}, \textit{gpt-oss-120b}, and \textit{gemma-4-31b-it}, reach 99.7\%, each answering 1,173 items correctly. Figure \ref{fig:acc-overall} presents each LLM's performance across question groups, question types, and IFC models.

Within the proprietary frontier tier, \textit{qwen3.7-max} reaches the benchmark ceiling, while \textit{gpt-5.6-luna} performs nearly perfectly. The smaller OpenAI model variants show lower performance, with \textit{gpt-5.4-mini} and \textit{gpt-5.4-nano} achieving 95.2\% and 85.8\%, respectively. The large open-weight tier exhibits a narrow performance range, from 97\% for \textit{llama-4-maverick} to 99.7\% for \textit{gpt-oss-120b}, with \textit{kimi-k3} (99.5\%) and \textit{deepseek-v4-flash} (99.2\%) also approaching the upper limit. The compact open-weight tier shows the greatest variation. At the upper end, \textit{qwen3.6-27b} matches the performance of the proprietary frontier models, while \textit{gemma-4-31b-it} (99.7\%) achieves the same accuracy level as \textit{gpt-5.6-luna} and \textit{gpt-oss-120b}. At the lower end, \textit{gemma-3-27b-it} achieves 87.4\%, substantially below its successor, and \textit{llama-3.1-8b-instruct}, the smallest model evaluated, fails to effectively execute the agent loop, achieving only 30.9\% accuracy, which is barely above the chance baseline.

\subsubsection{Performance by group and question type}
\label{sec:results-axes}

We observed differing performance amongst the evaluated LLMs across the four groups. Excluding the outlier performance of \textit{llama-3.1-8b-instruct}, direct retrieval items ask for a fact attached to a single entity and are answered almost perfectly (>97\%), except \textit{gpt-5.4-nano} (80.2\%) and \textit{gemma-3-27b-it} (86.2\%). Relational items require exploring relationships; all remaining models achieve at least 94\% in this group.  Quantitative items introduce counting, quantity lookup, aggregation, and comparison, while geometric items require AABB-based comparisons. These two groups prove more challenging, with noticeable performance degradation in \textit{gpt-5.4-mini}, which drops to 88.6\% on Quantitative and 88.8\% on Geometric, \textit{gpt-5.4-nano}, which drops to 86.2\% and 71.2\%, and \textit{gemma-3-27b-it}, which drops to 84.8\% and 68.8\%, respectively. Even among the highest-performing models, the few residual errors are concentrated primarily in these two groups, including \textit{deepseek-v4-pro}, \textit{minimax-m3}, \textit{deepseek-v4-flash}, and \textit{gemma-4-31b-it}.

By question type, single-choice items and boolean items are answered more accurately than multi-choice items. In multi-choice, the predicted answer set must contain every correct option and no distractor. That requirement exposes incomplete relational retrieval. Several mid-tier models that are strong on single-choice lose substantial accuracy on multi-choice, including \textit{llama-4-maverick} (83.8\%), \textit{nemotron-3-nano-30b-a3b} (88.6\%), \textit{gpt-oss-20b} (91.4\%), and \textit{gpt-5.4-mini} (91.4\%). By contrast, \textit{qwen3.7-max}, \textit{qwen3.6-27b}, \textit{deepseek-v4-pro}, \textit{deepseek-v4-flash}, and \textit{gemma-4-31b-it} all score 100\%.

\revision{Accuracy is similar across FZK Haus and Duplex, indicating that these reference IFC models present comparable difficulty under the tool agent setup. However, both models are relatively small and structurally simple, so this result does not establish that performance is invariant to the source IFC model. Stage 2 (Section \ref{tool-agent-additional-models}) therefore extends the evaluation to larger and more heterogeneous IFC models.}

\begin{figure}[t!]
    \centering
    \hspace*{-0.06\textheight}%
    \includegraphics[width=1.2\linewidth]{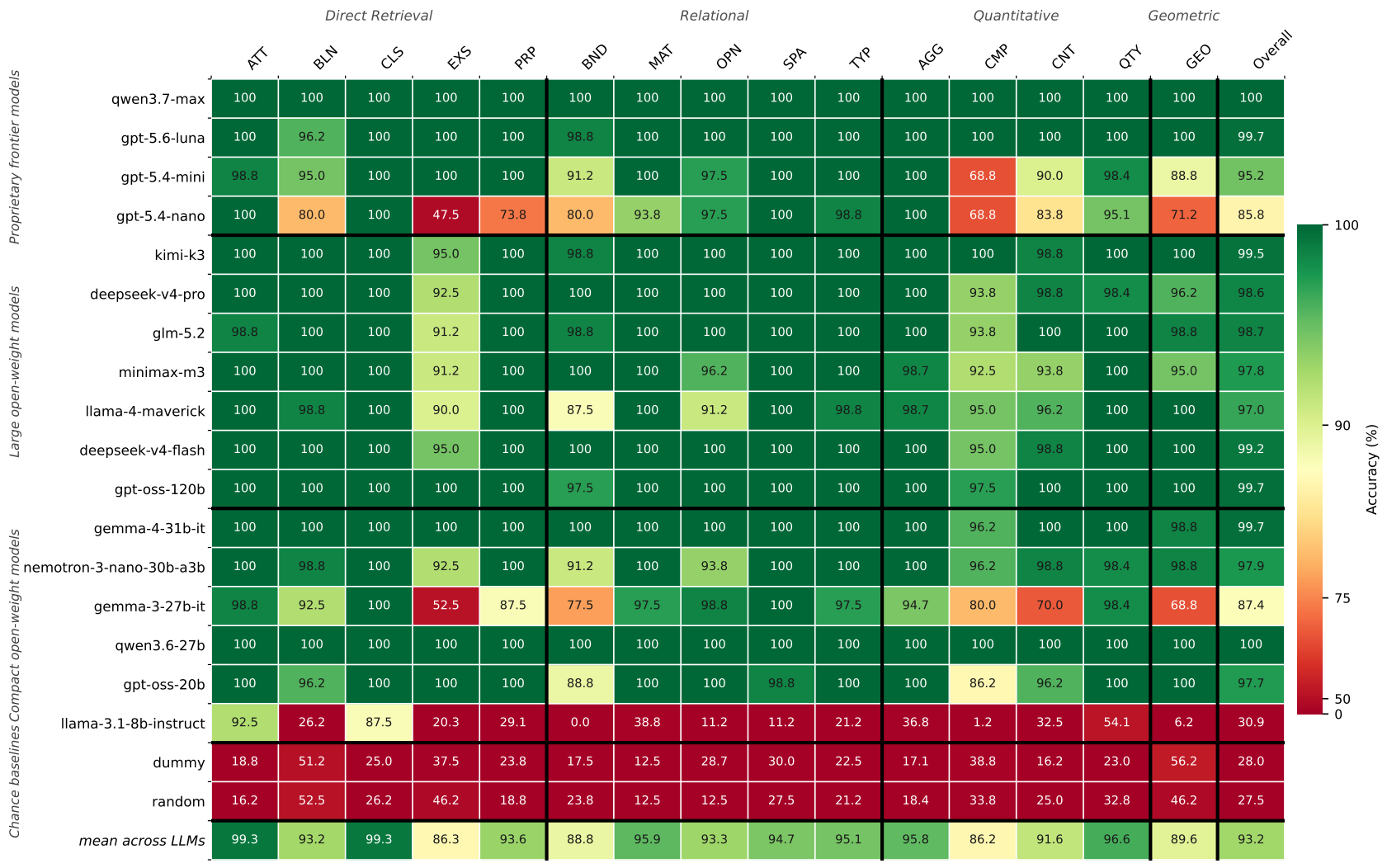}
    \caption{Tool agent accuracy by IFC-MCQ category for 17 LLMs on the 1,177-item Stage 1 benchmark.}
    \label{fig:acc-cat}
\end{figure}

\subsubsection{Category and Subtype-level Analysis}
\label{sec:results-subtype}
\begin{sidewaysfigure}[p]
    \centering
    \hspace*{-0.2\textheight}%
    \includegraphics[width=1.3\textheight]{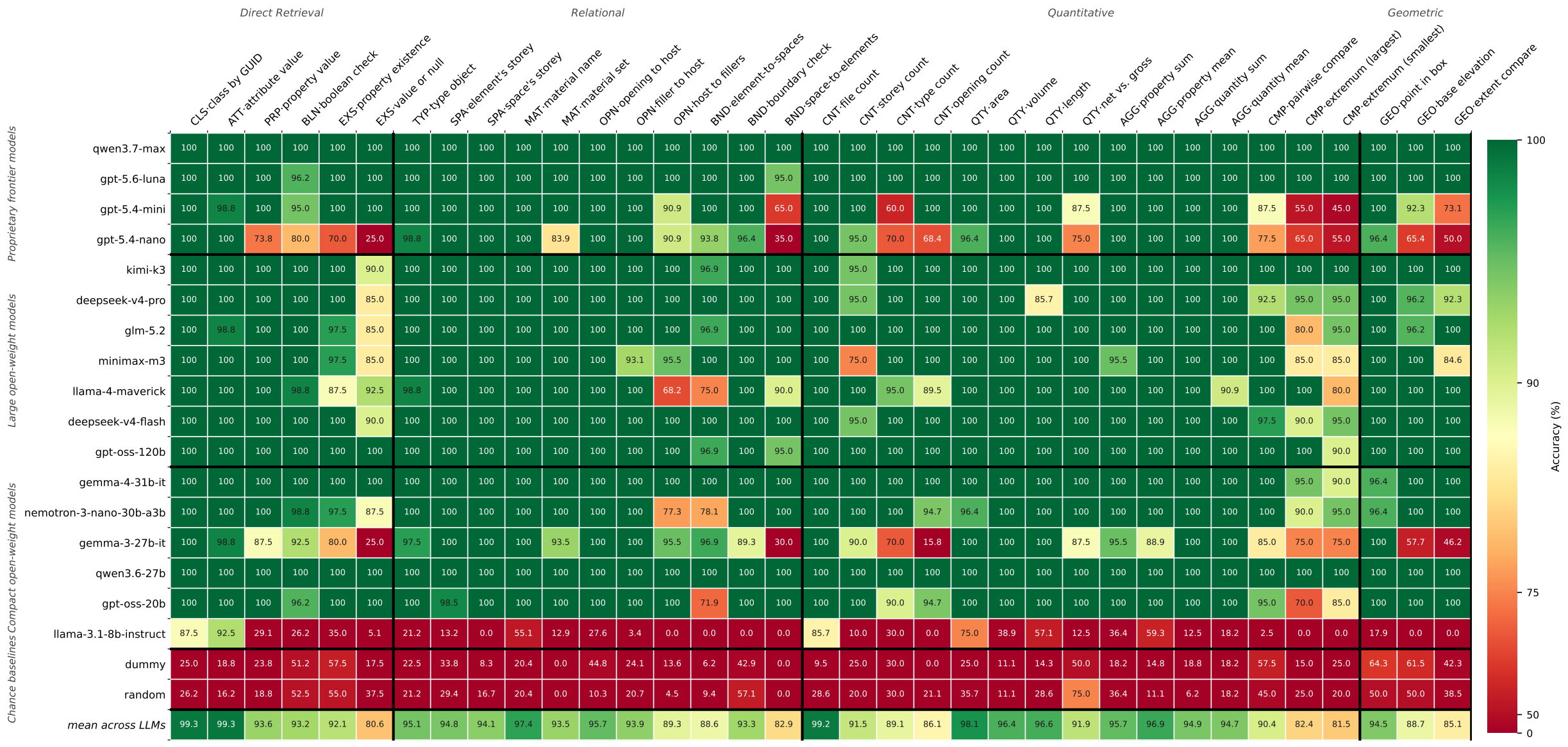}
    \caption{Tool agent accuracy by IFC-MCQ subtype for 17 LLMs on the 1,177-item Stage 1 benchmark.}
    \label{fig:acc-subtype}
\end{sidewaysfigure}

Figure \ref{fig:acc-cat} summarizes the performance of the evaluated LLMs across categories, and Figure \ref{fig:acc-subtype} across subtypes. Averaged across the 17 LLMs, the easiest categories are Attribute retrieval (ATT, 99.3\%), Class identification (CLS, 99.3\%), Quantity retrieval (QTY, 96.6\%), Material association (MAT, 95.9\%), and Aggregation (AGG, 95.8\%). These categories largely reduce to reading a stored value, a class label, a base quantity, a material name, or a tool-augmented sum or mean. The difficult categories are Existence / absence (EXS, 86.3\%), Comparison / extremum (CMP, 86.2\%), Space boundaries (BND, 88.8\%), and Geometry (GEO, 89.6\%), followed by Counting (CNT, 91.6\%). Boolean property verification (BLN, 93.2\%), Property retrieval (PRP, 93.6\%), and Openings and fillings (OPN, 93.3\%) are in between, while being easier for strong models.

Error counts across the 16 models other than \textit{llama-3.1-8b-instruct} exhibit the same overall ranking. Among the errors accumulated across these models, the largest numbers occur in Existence / absence (EXS, 122 errors), Comparison / extremum (CMP, 109), Space boundaries (BND, 72), and Geometry (GEO, 67). Examination of the dominant subtypes within these categories reveal the following observations:

\paragraph{Existence / absence (EXS) / value or null}
\label{sec:exs}
This subtype accounts for 94 of the 122 EXS errors. Items in this category ask for the value of a named property and list ``Not specified in the file'' as one of the options. Models that fail on these items typically return a plausible sibling colour, material, or dimension pulled from elsewhere in the file rather than acknowledging that the value is missing. This pattern is consistent with hallucination: the agent finds a nearby value and treats it as the answer instead of checking whether the queried property is actually populated on the queried entity. More capable reasoning models still miss several of these items, including \textit{deepseek-v4-pro/flash}, \textit{kimi-k3}, and \textit{minimax-m3}. In one example, a question asked whether the ``MainColor'' property was available for an \textit{IfcFurnishingElement}. \textit{qwen3.7-max} worked through the following tool sequence: \textit{get\_property\_sets(element)} > \textit{get\_type} > \textit{get\_type\_occurrences} > \textit{get\_property\_sets(type)} > \textit{get\_property\_values}, continuing until the absence of the property was explicitly confirmed. By contrast, \textit{deepseek-v4-pro} followed the same path only as far as \textit{get\_property\_sets(type)} before settling on a hallucinated colour value.

\paragraph{Comparison / extremum (CMP) / largest and smallest}
\label{sec:cmp}
Together these subtypes account for 83 of the 109 CMP errors. An extremum item names a shared metric and a candidate set of entities; the agent must read the metric for each candidate and select the extreme. Failures here are concentrated among \textit{gpt-5.4-mini/nano} in the proprietary tier, and several compact models, as well as large open-weights such as \textit{deepseek-v4-pro/flash}, \textit{glm-5.2}, and \textit{minimax-m3}. These errors primarily reflect difficulty maintaining conversational memory, incorrect tool use, and failure to follow the prescribed answering protocol. For instance, one question asked for the window with the smallest ``OverallHeight'' attribute. Models that answered correctly queried \textit{get\_entity} tool for every option and compared the returned values. Failures were more varied: some models chose tools that do not expose ``OverallHeight'', while others did not follow the one-action-per-turn protocol. For example, \textit{deepseek-v4-pro} checked the first few options, then stacked multiple actions, invented observations for the remaining windows, and answered before observing the asked attribute for every option.

\paragraph{Space boundaries (BND) / space-to-elements and element-to-spaces}
\label{sec:bnd}
These multi-choice subtypes dominate BND errors, with space-to-elements accumulating 38 errors and element-to-spaces 30. Recovering every bounding element of a space, or every space bounded by an element, requires using the \textit{get\_space\_boundaries} tool and selecting the correct answers from the provided options. These subtypes therefore require models to check the retrieved results carefully and discard incorrect answers. For space-to-elements, LLMs almost always do the right thing procedurally by calling the boundary tool; however, the failing models fail at multi-set matching. The dominant failure pattern was over-selection. For instance, \textit{gpt-5.4-mini} called \textit{get\_space\_boundaries} tool once on the target space and then answered with the gold set plus an extra option whose GUID never appeared in the tool's observation. For element-to-spaces, failures were more mixed: protocol violations, use of irrelevant tools, failure to produce a valid answer within the allotted turns, and over-selection. For instance, \textit{kimi-k3} called \textit{get\_space\_boundaries} tool on a relevant door and correctly observed a single bounded space. It then broke the one-action-per-turn protocol by stacking further calls and inventing observations while making assumptions such as a door should actually be bounding two spaces\footnote{The door in this example is the building's front entrance door.}, and eventually ended up selecting also a distractor space.

\paragraph{Geometry (GEO) / extent compare and base elevation}
\label{sec:geo}
Geometry errors are almost entirely confined to two subtypes: extent compare (40 errors) and base elevation (24 errors). The third subtype in this category, point-in-box, is comparatively easy once the bounding-box tool is called; the LLM then only needs to reason about whether the point lies within the returned box. All LLMs answered every question in this subtype correctly except for \textit{gpt-5.4-nano}, \textit{gemma-4-31b-it}, and \textit{nemotron-3-nano-30b-a3b}, each of which answered one question incorrectly.

The extent compare and base elevation subtypes, however, require reading two boxes, extracting the relevant scalar (axis extent or min-Z), and comparing them. In these subtypes, tool selection was not the issue; failures were due to extracting or reasoning over the box coordinates incorrectly. For example, on an extent compare item asking which of a door and a foundation wall has the larger extent in the Z dimension, \textit{gemma-3-27b-it} retrieved both AABBs via the \textit{get\_bounding\_box} tool and computed Z-extent correctly (door 2.108, wall 1.25), yet still selected the wall. On a base elevation item comparing a footing and a slab, the same model obtained min-Z values of -1.55 and 0.0 respectively, but answered that the footing sits higher. In both cases the numeric evidence needed was present in the observations; the error was in extracting and comparing the relevant coordinate. Notably, even the relatively small \textit{gpt-oss-20b} answered all questions across these subtypes correctly, whereas larger open-weight models such as \textit{deepseek-v4-pro} and \textit{glm-5.2} still struggled with a few questions.

\paragraph{Additional observations}
\label{sec:additional}
Within Counting (CNT), the opening count and type count subtypes account for 26 and 23 errors, respectively. In these subtypes, the agent must restrict the count to openings in a specific host or to occurrences of a named type. For opening count, LLMs usually call the \textit{get\_openings} tool correctly but then miscount the returned openings or confuse the host-specific count with the file-wide \textit{IfcOpeningElement} total. For type count, LLMs generally call the \textit{get\_type\_occurrences} tool correctly but then ignore the queried GUID and select the wrong count. Questions in the Spatial structure (SPA) and Type object resolution (TYP) categories are mostly answered correctly. Once an agent reliably calls \textit{get\_container} or \textit{get\_type}, these relational lookups appear largely solved for the models and files tested.

Particularly, \textit{gpt-5.4-nano} and \textit{gemma-3-27b-it} exhibit similar behaviour. These models perform well on questions that can be answered through direct lookups but struggle when the task requires reasoning, such as identifying missing information, performing multi-set matching, or correctly retrieving information from observations. Finally, the smallest evaluated LLM, \textit{llama-3.1-8b-instruct}, performs poorly across nearly every category, with only occasional correct answers in Attribute retrieval (ATT), Class identification (CLS), Counting (CNT), and Quantity retrieval (QTY).

\subsection{Stage 2: Tool agent on additional IFC models}
\label{tool-agent-additional-models}

\revision{As per the evaluation protocol discussed earlier in Section \ref{sec:evaluation-protocol}, Stage 2 retains the tool agent and evaluates five of the 17 LLMs, namely, \textit{gpt-5.6-luna}, \textit{deepseek-v4-flash}, \textit{gpt-oss-120b}, \textit{qwen3.6-27b}, and \textit{gemma-3-27b-it}, on four additional IFC models: Institute, Office, Trapelo, and TNW. Each of these additional models contributes 10 verified items per category to the evaluation benchmark. Consequently, FZK Haus, Duplex, Institute, Office, and Trapelo each contribute 150 items, whereas TNW contributes 140 items.}

\revision{On FZK Haus and Duplex items, the tool agent remains close to the Stage 1 result. \textit{gpt-5.6-luna} and \textit{qwen3.6-27b} answer every item. \textit{deepseek-v4-flash} answers 299 of 300 (99.7\%) and \textit{gpt-oss-120b} answers 298 of 300 (99.3\%). \textit{gemma-3-27b-it} answers 261 of 300 (87.0\%), in line with its 87.4\% on the full 1,177 items. However, on the additional IFC models introduced in this stage accuracy is lower for every LLM (refer to Table~\ref{tab:stage2-tools}). \textit{qwen3.6-27b} reaches 98.0\%, \textit{gpt-5.6-luna} 97.6\%, and \textit{gpt-oss-120b} 97.1\%. \textit{deepseek-v4-flash} falls from 99.7\% to 93.9\%. \textit{gemma-3-27b-it} falls from 87.0\% to 82.7\%. The lower accuracy is concentrated in a few categories. Across the five LLMs, Class identification (CLS), Attribute retrieval (ATT), and Spatial structure (SPA) questions are answered without error. The largest concentrations of errors occur in Aggregation (AGG; 43), Counting (CNT; 26), Existence / absence (EXS; 25), Comparison / extremum (CMP; 23), and Geometry (GEO; 20). Trapelo and TNW account for 117 of the 181 tool agent errors, and Institute and Office for 64.}

\begin{table}
[t!] \centering \footnotesize \caption{Tool agent overall performance with selected LLMs.} \label{tab:stage2-tools}
\begin{tabular}{lccc} 
\toprule \textbf{LLM} & \textbf{FZK Haus+Duplex} & \textbf{Institute+Office+Trapelo+TNW} \\ \midrule 
gpt-5.6-luna & 100.0 (300/300) & 97.6 (576/590) \\ 
deepseek-v4-flash & 99.7 (299/300) & 93.9 (554/590) \\ 
gpt-oss-120b & 99.3 (298/300) & 97.1 (573/590) \\ 
qwen3.6-27b & 100.0 (300/300) & 98.0 (578/590) \\ 
gemma-3-27b-it & 87.0 (261/300) & 82.7 (488/590) \\
\bottomrule 
\end{tabular} 
\end{table}

\revision{On the larger IFC models, a single tool call can return a long list, which can affect the performance of the agent. In this stage, \textit{deepseek-v4-flash} exhibited the largest performance degradation on the four additional IFC models. In one Trapelo item asking for the average height of 522 walls, \textit{get\_property\_values} returned the individual heights, but the output was truncated before the computed mean was available. This truncation resulted from the underlying implementation, which imposed a 6,000-character limit on tool outputs to prevent the retrieval of excessively long results. Consequently, the LLM could not see the computed sum or mean of all values, even though these were returned by the \textit{get\_property\_values} tool. The LLM correctly identified that the tool output had been truncated and therefore used the available values to estimate the average. However, \textit{deepseek-v4-flash} then assumed a unit scale and selected a nearby but incorrect value. This issue was not observed with \textit{gpt-5.6-luna}, \textit{qwen3.6-27b}, and \textit{gpt-oss-120b}. \textit{deepseek-v4-flash} also repeated the existence (EXS) error observed in Stage 1. When asked for ``MainColor'' on an \textit{IfcFurnishingElement} in Institute, where the property is absent, it answered a hallucinated colour. Its geometry failures were also similar to those cases observed in Stage 1.}

\revision{Overall, the reduced accuracy on the four additional IFC models appears to reflect a combination of category-specific difficulties and model-dependent retrieval conditions, including truncated tool outputs, rather than IFC model size alone.}

\subsection{Stage 3: Code agent versus tool agent}
\label{code-agent-all-models}

\revision{Stage 3 retains the same evaluation setup as Stage 2 but replaces the tool agent with the code agent. The overall results are summarized in Tables~\ref{tab:stage3} and \ref{tab:tool-code-comparison}. Compared with the tool agent scores, \textit{qwen3.6-27b}, \textit{gpt-oss-120b}, and \textit{gemma-3-27b-it} achieve lower accuracy with the code agent on both groups of IFC models. \textit{gpt-5.6-luna} answers five fewer items correctly on FZK Haus and Duplex, while achieving the same accuracy on the four additional models. \textit{deepseek-v4-flash} achieves lower accuracy on FZK Haus and Duplex but higher accuracy on the four additional models.}

\begin{table}
\centering
\footnotesize
\caption{Code agent overall performance with selected LLMs.}
\label{tab:stage3}
\begin{tabular}{lcc}
\toprule
\textbf{LLM}      & \textbf{FZK+Duplex} & \textbf{Institute+Office+Trapelo+TNW} \\ \midrule
gpt-5.6-luna      & 98.3 (295/300)      & 97.6 (576/590)                        \\
deepseek-v4-flash & 97.7 (293/300)      & 97.5 (575/590)                        \\

gpt-oss-120b      & 92.7 (278/300)      & 92.5 (546/590)                        \\
qwen3.6-27b       & 95.7 (287/300)      & 95.8 (565/590)                        \\
gemma-3-27b-it    & 69.7 (209/300)      & 62.7 (370/590)                        \\ \bottomrule
\end{tabular}
\end{table}

\begin{table}
\centering
\footnotesize
\caption{Performance comparison between tool and code agents.}
\label{tab:tool-code-comparison}
\begin{tabular}{lccc}
\toprule
\textbf{LLM} & \textbf{Tool agent, all six models} & \textbf{Code agent, all six models} & \textbf{Code $-$ tool} \\
\midrule
gpt-5.6-luna      & 98.4 (876/890) & 97.9 (871/890) & $-0.6$ pp \\
deepseek-v4-flash  & 95.8 (853/890) & 97.5 (868/890) & $+1.7$ pp \\
gpt-oss-120b       & 97.9 (871/890) & 92.6 (824/890) & $-5.3$ pp \\
qwen3.6-27b        & 98.7 (878/890) & 95.7 (852/890) & $-2.9$ pp \\
gemma-3-27b-it     & 84.2 (749/890) & 65.1 (579/890) & $-19.1$ pp \\
\bottomrule
\end{tabular}
\end{table}

\begin{table}
\centering
\footnotesize
\caption{Tool agent and code agent accuracy by IFC-MCQ category across six IFC models. Each non-BND category contains 300 answers, corresponding to 5 LLMs × 6 IFC models × 10 items. BND contains 250 answers because TNW has no BND items. After excluding \textit{gemma-3-27b-it}, the corresponding totals are 240 for non-BND categories and 200 for BND.}
\label{tab:stage3-category}
\begin{tabular}{@{}lcccc@{}}
\toprule
& \multicolumn{2}{c}{\textbf{Five LLMs}} & \multicolumn{2}{c}{\textbf{Other four LLMs, excluding \textit{gemma-3-27b-it}}} \\ \cmidrule(lr){2-3} \cmidrule(l){4-5} 
\textbf{Category}           & \textbf{Tool agent}     & \textbf{Code agent}     & \textbf{Tool agent}     & \textbf{Code agent}      \\ \midrule
CLS & 100.0 (300/300) & 100.0 (300/300) & 100.0 (240/240) & 100.0 (240/240) \\
ATT  & 100.0 (300/300) & 99.7 (299/300)  & 100.0 (240/240) & 100.0 (240/240) \\
PRP    & 95.7 (287/300)     & 98.3 (295/300)    & 100.0 (240/240)    & 99.2 (238/240)     \\
BLN      & 98.3 (295/300)     & 95.0 (285/300)    & 100.0 (240/240)    & 98.3 (236/240)     \\
EXS   & 87.7 (263/300)     & 92.0 (276/300)    & 97.5 (234/240)     & 90.8 (218/240)     \\
TYP       & 99.7 (299/300)     & 95.0 (285/300)    & 100.0 (240/240)    & 100.0 (240/240)    \\
SPA     & 100.0 (300/300)    & 86.3 (259/300)    & 100.0 (240/240)    & 95.4 (229/240)     \\
MAT  & 99.0 (297/300)     & 86.3 (259/300)    & 100.0 (240/240)    & 95.4 (229/240)     \\
OPN & 99.0 (297/300)     & 79.3 (238/300)    & 99.6 (239/240)     & 93.3 (224/240)     \\
BND      & 92.4 (231/250)     & 81.2 (203/250)    & 96.5 (193/200)     & 95.5 (191/200)     \\
CNT              & 89.3 (268/300)     & 91.3 (274/300)    & 94.6 (227/240)     & 97.5 (234/240)     \\
QTY    & 96.3 (289/300)     & 90.3 (271/300)    & 99.2 (238/240)     & 97.9 (235/240)     \\
AGG           & 85.7 (257/300)     & 87.0 (261/300)    & 87.5 (210/240)     & 95.0 (228/240)     \\
CMP & 90.7 (272/300)     & 90.7 (272/300)    & 94.6 (227/240)     & 99.6 (239/240)     \\
GEO              & 90.7 (272/300)     & 72.3 (217/300)    & 95.8 (230/240)     & 80.8 (194/240)     \\ \bottomrule
\end{tabular}
\end{table}

\revision{Table~\ref{tab:stage3-category} compares the two agents by category. Class identification (CLS) and Attribute retrieval (ATT) remain at or near 100\% accuracy with both agents. Across all five LLMs, the largest reductions with the code agent occur in Openings and fillings (OPN), Geometry (GEO), Spatial structure (SPA), Material association (MAT), and Space boundaries (BND). Moderate reductions are observed for Type resolution (TYP), Quantity retrieval (QTY), and Boolean property (BLN), whereas Property retrieval (PRP), Existence / absence (EXS), Counting (CNT), and Aggregation (AGG) show higher accuracy with the code agent, and Comparison / extremum (CMP) remains unchanged. However, several of these differences are strongly influenced by \textit{gemma-3-27b-it}. When this LLM is excluded, the differences between the two agents narrow substantially for SPA, MAT, OPN, and BND, while TYP reaches 100\% with both agents. Geometry remains the most pronounced difference, with code agent accuracy of 80.8\% compared with 95.8\% for the tool agent. Among the other four LLMs, the code agent also achieves reduced accuracy in PRP, BLN, EXS, and QTY, but improved accuracy in CNT, AGG, and CMP, reaching 97.5\%, 95.0\%, and 99.6\%, respectively.}

\revision{The two agents differ in how they retrieve the information required to answer a question. The tool agent selects a tool that directly returns the requested information and then interprets the resulting observation. In contrast, the code agent must retrieve the same information by generating and executing \textit{IfcOpenShell}-based code. The performance of \textit{gpt-5.6-luna} demonstrates that the two approaches can achieve comparable accuracy. However, the larger performance differences observed for the other LLMs indicate that high tool agent accuracy on FZK Haus and Duplex does not necessarily translate to comparable code agent accuracy, either on these models or on the additional IFC models.}

\subsection{No-IFC-access control}
\label{sec:no-ifc-control}

\revision{The no-IFC-access control described in Section~\ref{sec:evaluation-protocol} examines how much performance can be achieved from the question and answer options alone. Without access to the source IFC models, overall accuracy ranged from 37.0\% to 49.7\% across the four LLMs (Table~\ref{tab:no-ifc-control}). \textit{gpt-5.6-luna} and \textit{qwen3.6-27b} each achieved 49.7\%, \textit{gemma-3-27b-it} achieved 38.3\%, and \textit{llama-3.1-8b-instruct} achieved 37.0\%. Multi-choice items produced the lowest accuracy for every model. Although boolean accuracy was higher for all four models, boolean items contain only two options and therefore have a 50\% random-selection baseline.}

\revision{For the three LLMs also evaluated in Stages 2 and 3, access to the IFC models substantially increased accuracy. Accuracy increased from 49.7\% to 98.4\% and 97.9\% for \textit{gpt-5.6-luna} with the tool and code agents, respectively; from 49.7\% to 98.7\% and 95.7\% for \textit{qwen3.6-27b}; and from 38.3\% to 84.2\% and 65.1\% for \textit{gemma-3-27b-it}. This is not a pure ablation of IFC access because the agentic configurations also provide a retrieval interface and iterative interaction. Nevertheless, the results show that information available in the question and options alone does not account for the substantially higher accuracy observed under the agentic configurations.}

\begin{table}
\centering
\footnotesize
\caption{Accuracy without access to the source IFC models on the 890 items used in Stages 2 and 3.}
\label{tab:no-ifc-control}
\begin{tabular}{@{}lrrrr@{}}
\toprule
\textbf{LLM} & \textbf{Overall} & \textbf{Single} &
\textbf{Multi} & \textbf{Boolean} \\
\midrule
\multicolumn{4}{l}{\textit{high-performing proprietary and open-weight LLMs in Stage 2 and 3}} \\
gpt-5.6-luna
& 49.7 (442/890) & 49.7 & 30.1 & 61.5 \\
qwen3.6-27b
& 49.7 (442/890) & 49.6 & 24.7 & 65.8 \\ \midrule
\multicolumn{4}{l}{\textit{low-performing LLM in Stages 2 and 3}} \\
gemma-3-27b-it
& 38.3 (341/890) & 38.1 & 8.2 & 58.1 \\ \midrule
\multicolumn{3}{l}{\textit{low-performing LLM in Stage 1}} \\
llama-3.1-8b-instruct
& 37.0 (329/890) & 39.0 & 6.8 & 43.6 \\
\bottomrule
\end{tabular}
\end{table}

\section{Discussion}
\label{sec:discussion}
In this section, we first interpret agent interaction patterns and the role of the retrieval interface. We then discuss the implications of our findings, followed by the limitations and opportunities for future extensions.

\subsection{Agent interaction and resource usage}
\label{sec:agent-behaviour}
This subsection examines how the agent configurations differ in their interaction steps, observation truncation, and token usage.

\paragraph{Interaction steps and observation truncation}
In Stage 1, on FZK Haus and Duplex, the tool agent answers most items with a single tool call. For LLMs achieving above 95\% accuracy, the mean ranges from 1.15 to 1.33 calls per item, with 82--88\% of items answered after a single call. This pattern is consistent with the design of the tools, as many direct and simple relational questions can be resolved using a single correctly selected tool. \textit{llama-3.1-8b-instruct} is the exception: it averages 6.01 calls per item, answers only 18\% of items after a single call, and achieves 30.9\% accuracy.

\revision{In Stages 2 and 3 (results summarized in Table~\ref{tab:calls-six-models}), on FZK Haus and Duplex, the tool agent answers 82.7\% and 87.9\% of items, respectively, after a single call, consistent with the pattern observed in Stage 1. On the four additional IFC models, it continues to answer most items with a single call, ranging from 78.0\% on Trapelo to 84.0\% on Office. In contrast, the code agent (which calls the \textit{run\_python} tool; refer to Figure \ref{fig:code-agent-system-prompt}) more frequently requires additional calls. Its mean ranges from 1.77 calls per item on Duplex to 2.09 on Institute, while the proportion of items answered after a single call ranges from 60.9\% on Institute to 66.6\% on TNW. Tool observations were truncated at 6,000 characters. The tool agent reaches this limit for 186 answers, all on the four additional IFC models, including 82 on TNW. The code agent reaches the limit for 81 answers across the six IFC models.}

\begin{table} 
\centering 
\footnotesize \caption{Calls per item across the six IFC models for the five LLMs in Stages 2 and 3. \textit{One call} denotes the percentage of items answered after a single call. \textit{Truncated} denotes the number of answers for which an observation was truncated at 6,000 characters.} 
\label{tab:calls-six-models} 
\begin{tabular}{@{}l cc cc cc@{}} \toprule & \multicolumn{2}{c}{\textbf{Mean calls}} & \multicolumn{2}{c}{\textbf{One call (\%)}} & \multicolumn{2}{c}{\textbf{Truncated}} \\
\cmidrule(lr){2-3}\cmidrule(lr){4-5}\cmidrule(lr){6-7} \textbf{IFC model} & \textbf{Tool} & \textbf{Code} & \textbf{Tool} & \textbf{Code} & \textbf{Tool} & \textbf{Code} \\
\midrule FZK Haus & 1.30 & 1.89 & 82.7 & 63.7 & 0 & 1 \\
Duplex & 1.17 & 1.77 & 87.9 & 63.7 & 0 & 9 \\ \midrule
Institute & 1.34 & 2.09 & 80.3 & 60.9 & 39 & 15 \\
Office & 1.22 & 1.92 & 84.0 & 62.0 & 26 & 7 \\
Trapelo & 1.36 & 1.91 & 78.0 & 61.3 & 39 & 12 \\
TNW & 1.32 & 1.79 & 80.9 & 66.6 & 82 & 37 \\
\bottomrule 
\end{tabular} 
\end{table}

\paragraph{Token usage}

\revision{The mean input and output tokens per item for LLMs evaluated in Stages 2 and 3 are reported in Table~\ref{tab:tokens}. With the tool agent, mean input ranges from 2,533 tokens per item for \textit{gpt-5.6-luna} to 5,369 for \textit{deepseek-v4-flash}, while their mean output ranges from 98 to 1,904 tokens per item. The code agent uses more input tokens than the tool agent for every LLM except \textit{gpt-5.6-luna}. The code agent also uses more output tokens for every LLM. \textit{deepseek-v4-flash} has the highest token usage with the code agent, averaging 9,480 input and 3,276 output tokens per item.}

\begin{table}
\centering 
\footnotesize 
\caption{Mean input and output tokens per item across the six IFC models for the five LLMs in Stages 2 and 3. Output includes reasoning tokens, if applicable.} 
\label{tab:tokens} 
\begin{tabular}{@{}l cc cc@{}} 
\toprule & \multicolumn{2}{c}{\textbf{Tool}} & \multicolumn{2}{c}{\textbf{Code}} \\
\cmidrule(lr){2-3}\cmidrule(lr){4-5} \textbf{LLM} & \textbf{Input} & \textbf{Output} & \textbf{Input} & \textbf{Output} \\
\midrule 
gpt-5.6-luna & 2,533 & 98 & 2,398 & 188 \\
qwen3.6-27b & 2,667 & 1107 & 5,000 & 1,476 \\
gpt-oss-120b & 3,716 & 347 & 4,247 & 948 \\
deepseek-v4-flash & 5,369 & 1,904 & 9,480 & 3,276 \\
gemma-3-27b-it & 4,213 & 200 & 5,302 & 552 \\
\bottomrule 
\end{tabular} 
\end{table}

\subsection{Role of the retrieval interface}
\revision{The high scores achieved by several LLMs with the tool agent should be interpreted in relation to the retrieval interface. The 17 tools were designed around the current IFC-MCQ taxonomy and provide at least one intended retrieval route for every category. For many items, an LLM can answer by selecting the relevant tool, providing the required arguments, reading its output, and constructing the answer set. Across the six IFC models in Stage 2, 78.0--87.9\% of tool agent answers were completed after one tool call, depending on the source model.}

\revision{The code agent results show that performance also varies with the agent design. The code agent does not have access to any predefined IFC retrieval tools and must instead identify the relevant IFC structures, write \textit{IfcOpenShell}-based Python code, execute it through the \textit{run\_python} tool, and interpret the result. Several LLMs achieved reduced overall and category-level accuracy with the code agent, particularly for spatial relationships, materials, openings and fillings, space boundaries, and geometry. However, \textit{deepseek-v4-flash} improved under code execution, showing that the code interface is not consistently more difficult for every LLM. For most evaluated LLMs, the code agent used more input and output tokens. The exception was \textit{gpt-5.6-luna} for input tokens, whereas output-token usage was higher for every LLM. The two interfaces therefore place different demands on the evaluated systems.}

\revision{Overall, these results show that the performance observed in the agentic evaluation varies across LLMs and agent configurations. The near-ceiling accuracies observed with the tool agent should therefore be interpreted in the context of its predefined, taxonomy-complete tool set, which makes many retrieval tasks relatively direct.}

\subsection{Implications for benchmark generation and agentic evaluation}
\label{sec:implications}
\revision{The findings have two distinct sets of implications. Implications (i) and (ii) concern benchmark generation, verification, and category coverage across source IFC models. Implications (iii) and (iv) concern the performance of the evaluated LLM-agent configurations.}

\revision{(i) \textit{IFC-MCQ provides a basis for evaluating IFC data retrieval without relying on a human or LLM judge.} As described in Section~\ref{sec:method-pipeline} and illustrated in Figure~\ref{fig:method}, each category generator builds a candidate pool from facts represented in the model index, fills in a stem template, constructs distractors, and stores a metadata record that provides the interface to the separate verification path. The experiments (Section \ref{sec:results}) demonstrate that benchmark items can be generated directly from IFC files, re-derived through a separate verification path, and graded using exact option-set matching. All items retained in the benchmark passed verification, while a small number of candidates from TNW, Trapelo, and Sample House were rejected because the separately derived answer did not agree with the generated answer (Table \ref{tab:generation-results}). A stratified human audit provided complementary supporting evidence: all sampled items were grammatically well formed, their independently retrieved answers matched the benchmark keys, and those answers were uniquely represented among the options. Together, these properties support the use of IFC-MCQ for repeatable comparison and regression testing without introducing variability from subjective answer assessment.}

\revision{(ii) \textit{The source IFC model affects the category coverage of a generated benchmark.} Applying IFC-MCQ to ten IFC models (Section \ref{sec:generation-results}) showed that the number and types of questions that can be generated depend on the information represented in each model. A category may not yield any items when the required IFC relationships are absent, while candidate-generation attempts may be skipped when too few distinct values are available to construct a valid option-set. For example, TNW and UNSW Block House produced no Space boundaries (BND) items because these files did not contain \textit{IfcRelSpaceBoundary} relationships, and all 402 candidate-generation skips occurred in Property retrieval (PRP) or Existence / absence (EXS). Reporting the characteristics of the source IFC models and the resulting category coverage is therefore important when interpreting the benchmark-generation results. Specific opportunities to extend IFC-MCQ components are discussed later in Section~\ref{sec:release}.}

\revision{(iii) \textit{The retrieval interface affects the measured performance of an LLM-based IFC querying system.} Using the same five LLMs and six IFC models, replacing the fixed tool interface with code execution changed both overall and category-level accuracy. The magnitude and direction of these differences varied across LLMs and question categories. For the tasks evaluated in this study, the tool agent is the more suitable configuration when the predefined tools provide an intended retrieval route for the required category, as it generally achieved higher accuracy and required fewer interaction steps and lower token usage. The code agent provides a less constrained retrieval interface, but its performance on tasks beyond the current IFC-MCQ taxonomy was not evaluated. Moreover, \textit{deepseek-v4-flash} improved under code execution, showing that one interface was not consistently better for every LLM. LLM-based IFC querying systems should therefore be evaluated using the retrieval interface intended for deployment, and the reported performance should be interpreted at the level of the complete LLM-agent configuration.}

\revision{(iv) \textit{Overall accuracy can conceal category-specific weaknesses.} Class identification and attribute retrieval were consistently answered with high accuracy, whereas larger error concentrations occurred in tasks involving missing information, relationship retrieval, counting and aggregation, comparison, and geometry. These differences remain visible even when overall accuracy is high. Reporting results by category and, where sample sizes permit, by subtype therefore provides a more informative assessment of where an IFC querying system succeeds or fails than overall accuracy alone.}

\subsection{Limitations}
\label{sec:limitations}

\revision{This study has several limitations. First, IFC-MCQ evaluates retrieval through multiple-choice questions rather than free-text responses. Although this enables deterministic grading without human or LLM judges, performance may depend on the available options and distractor construction, and the template-based questions do not capture the full linguistic variation, ambiguity, or conversational context of real user queries. In the no-IFC-access control, selected LLMs answered 37.0--49.7\% of the items correctly using only the question and options. This performance may reflect prior model knowledge, cues in the questions or options, option elimination, or chance. IFC-MCQ scores should therefore be interpreted within the generated multiple-choice setting rather than as a direct measure of retrieval alone or of open-ended question-answering performance. In addition, the near-ceiling accuracies observed with the tool agent limit fine-grained comparison among the highest-performing systems in that configuration.}

\revision{Second, although candidate answers are re-derived through a separate verification path, both generation and verification rely on \textit{IfcOpenShell}. The verification stage can detect disagreements between the two implementations but cannot eliminate errors arising from behaviour shared through the underlying IFC-processing library. The current geometric category is also restricted to operations based on axis-aligned bounding boxes and does not evaluate more general solid geometry, topology, or spatial reasoning. In addition, the human audit covered a stratified sample of items across six IFC models and therefore provides supporting evidence for the sampled items rather than validation of every retained item. Further, the implementation was tested on ten parseable IFC files and does not evaluate malformed, corrupted, or otherwise non-parseable IFC files. Such cases may prevent model indexing or affect the availability of properties and relationships, and robustness to these conditions remains outside the scope of the current study.}

\revision{Finally, observation truncation affected a subset of answers from both the tool and code agents. In these cases, the agents may not have received all retrieved evidence. Performance on queries that produced large outputs may therefore have been affected by incomplete observations.}

\subsection{Software availability and extensibility}
\label{sec:release}

\revision{IFC-MCQ is implemented as a modular Python library and will be made available under the Apache 2.0 License through the Python Package Index (PyPI) upon acceptance.}

\revision{The modular design also provides specific directions for extending the current pipeline. Candidate pools can be broadened by adding further lookups to the model index and new category generators for additional IFC mechanisms and multi-step retrieval tasks. The fixed stem-template pools can be expanded with additional wordings or structures to increase linguistic variation within each subtype. Distractor construction can be refined through category-specific strategies for source IFC models with too few distinct alternative values. Metadata records can likewise be extended to represent the additional entities and relationships required by more complex items and to provide the information needed by corresponding verifiers.}

\section{Conclusion}
\label{sec:conclusion}

\revision{This paper introduced IFC-MCQ, a method for automatically generating MCQ benchmarks from IFC files for evaluating LLM-based IFC data retrieval. Its taxonomy comprises 15 categories and 36 subtypes organized into four groups: Direct retrieval, Relational, Quantitative, and Geometric. Each category generator constructs question stems, answer key(s), and distractors from the input IFC file. Before a candidate item is retained, its answer is re-derived through a separate verification path that does not receive the stored answer key. Retained items can therefore be scored using exact option-set matching without requiring a human or LLM judge.}

\revision{We applied IFC-MCQ to ten IFC models of varying sizes and modelling characteristics. Benchmark coverage varied because some models lacked the relationships, quantities, or distinct values required by particular categories. Most candidate items passed verification, while a small number were rejected because the generated and re-derived answers disagreed. A stratified human audit provided complementary supporting evidence for the sampled items: all were grammatically well formed, their independently retrieved answers matched the benchmark answer keys, and those answers were uniquely represented among the options. These results indicate that the source IFC model affects benchmark coverage and can expose generation and verification edge cases.}

\revision{The three-stage agentic evaluation examined LLM performance, variation across source IFC models, and the effect of the retrieval interface. Stage 1 evaluated 17 LLMs with the tool agent on the reference models FZK Haus and Duplex. Stages 2 and 3 evaluated a selected set of five LLMs on these models and four additional models using the tool and code agents, respectively. In Stage 1, most evaluated LLMs achieved high overall accuracy, although category-level results revealed remaining weaknesses. In Stage 2, tool-agent accuracy was generally lower on the four additional IFC models, with errors concentrated in a subset of categories. In Stage 3, replacing the fixed tool interface with code execution changed both overall and category-level accuracy, with the magnitude and direction of these differences varying across LLMs and question categories. The code agent also generally required more interaction steps and tokens. Overall, the results show that observed performance can vary across source IFC models and LLM-agent configurations and should therefore be interpreted in the context of both the source model and retrieval interface.}

Beyond the benchmark-generation method itself, this work contributes to construction automation in three practical ways. First, the taxonomy of 15 categories and 36 subtypes provides a systematic basis for evaluating natural language access to IFC data and identifying the types of building-information retrieval on which a system performs reliably within the benchmark. Second, the category-level analysis reveals error patterns relevant to construction applications, including selecting values for absent properties, incomplete relationship retrieval, incorrect counting or aggregation, and errors in geometric comparison. Such patterns may be obscured by high overall accuracy but can be important when retrieved building information supports engineering, coordination, or facility-management tasks. Third, the expanded experiments indicate that evaluation on a small number of reference models or with a single retrieval interface may not characterise performance under other configurations. This supports evaluating LLM-based IFC querying systems on project models representative of their intended application and using the retrieval interface intended for deployment.

\revision{The findings should be interpreted within the limitations (Section \ref{sec:limitations}) of the current method. IFC-MCQ evaluates a multiple-choice setting in which the questions and answer options may provide cues; geometric coverage is limited to AABB-based operations; generation and verification share the underlying IFC-processing library; and the reported performance is specific to the evaluated LLMs, source IFC models, and retrieval interfaces. Future work will extend the taxonomy to more general geometric and multi-step retrieval tasks, apply the pipeline to additional IFC models, evaluate further agent architectures and retrieval interfaces, and examine whether performance in the multiple-choice setting transfers to free-text IFC question-answering.}

\section*{Acknowledgments}
This work was supported by the Australian Research Council (ARC) Industry Transformation Research Hub (ITRH) for Resilient and Intelligent Infrastructure Systems (RIIS) under Grant: ARC ITRP IH210100048.

\section*{Declaration of AI and AI-assisted Tools in the manuscript preparation process}
During the preparation of this manuscript, the authors used ChatGPT and Claude to assist with grammar and language editing.

\bibliographystyle{elsarticle-num-names}
\bibliography{ref}

@article{lin2016natural,
  title={A natural-language-based approach to intelligent data retrieval and representation for cloud {BIM}},
  author={Lin, Jia-Rui and Hu, Zhen-Zhong and Zhang, Jian-Ping and Yu, Fang-Qiang},
  journal={Computer-Aided Civil and Infrastructure Engineering},
  volume={31},
  number={1},
  pages={18--33},
  year={2016},
  publisher={Wiley Online Library},
  doi={10.1111/mice.12151}
}

@article{shin2021bimasr,
  title={{BIMASR}: Framework for voice-based {BIM} information retrieval},
  author={Shin, Sangyun and Issa, Raja RA},
  journal={Journal of Construction Engineering and Management},
  volume={147},
  number={10},
  pages={04021124},
  year={2021},
  publisher={American Society of Civil Engineers},
  doi={10.1061/(ASCE)CO.1943-7862.0002138}
}

@article{yin2023two,
  title={Two-stage {Text-to-BIMQL} semantic parsing for building information model extraction using graph neural networks},
  author={Yin, Mengtian and Tang, Llewellyn and Webster, Chris and Li, Jinyang and Li, Haotian and Wu, Zhuoquan and Cheng, Reynold CK},
  journal={Automation in construction},
  volume={152},
  pages={104902},
  year={2023},
  publisher={Elsevier},
  doi={10.1016/j.autcon.2023.104902}
}

@article{ibba2026ask,
  title={{ASK-BIM}: A knowledge graph-powered {AI} system for natural language querying of {BIM} models},
  author={Ibba, Andrea and Alonso, Rub{\'e}n and Recupero, Diego Reforgiato},
  journal={Data \& Knowledge Engineering},
  volume={164},
  pages={102581},
  year={2026},
  publisher={Elsevier},
  doi={10.1016/j.datak.2026.102581}
}

@online{pacheco2024bimconverse,
  author    = {Pacheco, Libny and Berkmiller, Christoph},
  title     = {{BIMConverse} – {GraphRAG} for {IFC} Natural Language Queries},
  year      = {2024},
  url       = {https://blog.iaac.net/bimconverse-graphrag-for-ifc-natural-language-queries/},
  note      = {Accessed 2026-08-10}
}

@article{lamsal2026ifcllm,
  title={{IfcLLM}: Natural Language Querying of {IFC} Models through Complementary Relational and Graph Representations},
  author={Lamsal, Rabindra and Zlatanova, Sisi and Xu, Haowen and Sun, Yafei and Shen, Johnson Xuesong},
  journal={arXiv preprint arXiv:2605.13236},
  year={2026},
  doi={10.48550/arXiv.2605.13236}
}

@inproceedings{iranmanesh2025llm,
  title={{LLM}-assisted {Graph-RAG} information extraction from {IFC} data},
  author={Iranmanesh, Sima and Saadany, Hadeel and Vakaj, Edlira},
  booktitle={European Conference on Computing in Construction},
  year={2025},
  doi = {10.35490/EC3.2025.366},
}

@article{liu2025bimcoder,
title={{BIMCoder}: A comprehensive large language model fusion framework for natural language-based {BIM} information retrieval},
  author={Liu, Bingru and Chen, Hainan},
  journal={Applied Sciences},
  volume={15},
  number={14},
  pages={7647},
  year={2025},
  publisher={MDPI},
  doi={10.3390/app15147647}
}

@inproceedings{ifcbenchv1,
  title = {Natural Language Information Retrieval from {BIM} Models: An {LLM}-Based Multi-Agent System Approach},
  booktitle = {European Conference on Computing in Construction},
  author = {Hellin, Sylvain and Nousias, Stavros and Borrmann, Andr{\'e}},
  year = 2025,
  doi = {10.35490/EC3.2025.265},
}

@article{ifcbenchv2,
  title={Agentic search for {BIM} information extraction: Systematic evaluation},
  author={Hellin, Sylvain and Jang, Suhyung and Fuchs, Stefan and Nousias, Stavros and Borrmann, Andr{\'e}},
  journal={Automation in Construction},
  volume={192},
  pages={107260},
  year={2026},
  publisher={Elsevier},
  doi={10.1016/j.autcon.2026.107260}
}

@mastersthesis{avgoren2025enhancing,
  title={Enhancing {IFC} Model Interpretability Using Knowledge Graphs and Large Language Models with Integrated Visual Support},
  author={Avgoren, Atacan Kural},
  school = {Technical University of Munich},
  year={2025},
  url={https://mediatum.ub.tum.de/node?id=1782062}
}

@inproceedings{volkova2026ifc,
  title={{IFC-QA}: A Benchmark for Evaluating {LLM} Reliability over Industrial Building Information Model},
  author={Volkova, Anastasia and Gromyko, Andrew and Gusarov, Anton and Maslov, Evgenii and Khrulkov, Valentin and Kuznetsov, Andrey and Oseledets, Ivan},
  booktitle={AI4X - Accelerate Conference 2026},
  year={2026},
  url = {https://openreview.net/forum?id=H6vTDjJMZ9}
}

@inproceedings{duifcmemorybench,
  title={{IFCMemoryBench}: Evaluating Long-Term Memory of {LLM}-Based Agents in {BIM} Information Retrieval},
  author={Du, Changyu and Vosseler, Alexander and Mazza, Filippo and Borrmann, Andr{\'e}},
  booktitle={KDD workshop on Evaluation and Trustworthiness of Agentic AI},
  year={2026},
  doi={10.48550/arXiv.2607.26072}
}

@article{clark2018think,
  title={Think you have solved question answering? try {ARC}, the {AI2} reasoning challenge},
  author={Clark, Peter and Cowhey, Isaac and Etzioni, Oren and Khot, Tushar and Sabharwal, Ashish and Schoenick, Carissa and Tafjord, Oyvind},
  journal={arXiv preprint arXiv:1803.05457},
  year={2018},
  doi={10.48550/arXiv.1803.05457}
}

@inproceedings{hendrycks2020measuring,
  title={Measuring massive multitask language understanding},
  author={Hendrycks, Dan and Burns, Collin and Basart, Steven and Zou, Andy and Mazeika, Mantas and Song, Dawn and Steinhardt, Jacob},
  booktitle={International Conference on Learning Representations},
  year={2021},
  url={https://openreview.net/forum?id=d7KBjmI3GmQ}
}

@inproceedings{rein2023gpqa,
title={GPQA: A graduate-level google-proof Q\&A benchmark},
  author={Rein, David and Hou, Betty Li and Stickland, Asa Cooper and Petty, Jackson and Pang, Richard Yuanzhe and Dirani, Julien and Michael, Julian and Bowman, Samuel R},
  booktitle={First Conference on Language Modeling},
  year={2024},
  url={https://openreview.net/forum?id=Ti67584b98}
}

@article{wang2019integration,
  title={Integration of {BIM} and {GIS} in sustainable built environment: A review and bibliometric analysis},
  author={Wang, Hao and Pan, Yisha and Luo, Xiaochun},
  journal={Automation in construction},
  volume={103},
  pages={41--52},
  year={2019},
  publisher={Elsevier},
  doi={10.1016/j.autcon.2019.03.005}
}

@book{eastman2011bim,
  title={{BIM} handbook: A guide to building information modeling for owners, managers, designers, engineers and contractors},
  author={Eastman, Chunk and Teicholz, Paul and Sacks, Rafael and Liston, Kathleen},
  year={2011},
  publisher={John Wiley \& Sons},
  isbn={978-0-470-54137-1}
}

@misc{ifc,
  author       = {{buildingSMART International}},
  title        = {Industry Foundation Classes ({IFC})},
  howpublished = {\url{https://technical.buildingsmart.org/standards/ifc/}},
  year         = {{2025}},
  note={Accessed 2026-08-10}
}

@article{achiam2023gpt,
  title={{GPT-4} technical report},
  author={Achiam, Josh and Adler, Steven and Agarwal, Sandhini and Ahmad, Lama and Akkaya, Ilge and Aleman, Florencia Leoni and Almeida, Diogo and Altenschmidt, Janko and Altman, Sam and Anadkat, Shyamal and others},
  journal={arXiv preprint arXiv:2303.08774},
  year={2023},
  doi={10.48550/arXiv.2303.08774}
}

@article{zhao2023survey,
  title={A survey of large language models},
  author={Zhao, Wayne Xin and Zhou, Kun and Li, Junyi and Tang, Tianyi and Dong, Zican and Hou, Yupeng and Zhang, Beichen and Min, Yingqian and Zhang, Junjie and Liu, Peiyu and others},
  journal={Frontiers of Computer Science},
  volume={20},
  number={12},
  pages={2012627},
  year={2026},
  publisher={Springer},
  doi={10.1007/s11704-026-60308-3}
}

@article{cai2025survey,
  title={A survey on mixture of experts in large language models},
  author={Cai, Weilin and Jiang, Juyong and Wang, Fan and Tang, Jing and Kim, Sunghun and Huang, Jiayi},
  journal={IEEE Transactions on Knowledge and Data Engineering},
  volume={37},
  number={7},
  pages={3896--3915},
  year={2025},
  publisher={IEEE},
  doi={10.1109/TKDE.2025.3554028}
}

@misc{qwen37,
    title = {{Qwen3.7}: The Agent Frontier},
    url = {https://qwen.ai/blog?id=qwen3.7},
    author = {{Qwen Team}},
    howpublished = {Qwen Blog},
    month = {May},
    year = {2026},
    note = {Accessed 2026-08-10}
}

@misc{openai2026gpt56,
  title        = {{GPT-5.6}: Frontier Intelligence that Scales with Your Ambition},
  author       = {{OpenAI}},
  year         = {2026},
  month        = jul,
  howpublished = {OpenAI Blog},
  url          = {https://openai.com/index/gpt-5-6/},
  note         = {Accessed 2026-08-10}
}

@misc{openai2026gpt54mininano,
  title        = {Introducing {GPT-5.4} Mini and Nano},
  author       = {{OpenAI}},
  year         = {2026},
  month        = mar,
  howpublished = {OpenAI Blog},
  url          = {https://openai.com/index/introducing-gpt-5-4-mini-and-nano/},
  note         = {Accessed 2026-08-10}
}

@article{team2026kimi,
  title={Kimi {K3}: Open Frontier Intelligence},
  author={{Kimi Team}},
  journal={arXiv preprint arXiv:2607.24653},
  year={2026},
  doi={10.48550/arXiv.2607.24653}
}

@article{xu2026deepseek,
  title={{DeepSeek-V4}: Towards highly efficient million-token context intelligence},
  author={Xu, Anyi and Lin, Bangcai and Xue, Bing and Wang, Bingxuan and Xu, Bingzheng and Wu, Bochao and Zhang, Bowei and Lin, Chaofan and Dong, Chen and Ling, Chenchen and others},
  journal={arXiv preprint arXiv:2606.19348},
  year={2026},
  doi={10.48550/arXiv.2606.19348}
}

@article{zeng2026glm,
  title={{GLM-5}: from vibe coding to agentic engineering},
  author={Zeng, Aohan and Lv, Xin and Hou, Zhenyu and Du, Zhengxiao and Zheng, Qinkai and Chen, Bin and Yin, Da and Ge, Chendi and Huang, Chenghua and Xie, Chengxing and others},
  journal={arXiv preprint arXiv:2602.15763},
  year={2026},
  doi={10.48550/arXiv.2602.15763}
}

@article{lai2026minimax,
  title={{MiniMax} sparse attention},
  author={Lai, Xunhao and Xu, Weiqi and Yang, Yufeng and Chen, Qiaorui and Xu, Yang and Zeng, Lunbin and Li, Xiaolong and Sun, Haohai and Zhu, Haichao and Zhang, Vito and others},
  journal={arXiv preprint arXiv:2606.13392},
  year={2026},
  doi={10.48550/arXiv.2606.13392}
}

@article{meta2025llama,
  title={The {Llama 4} herd: The beginning of a new era of natively multimodal {AI} innovation},
  author={Meta},
  journal={AI at Meta},
  year={2025},
  url={https://ai.meta.com/blog/llama-4-multimodal-intelligence/},
  note={Accessed 2026-08-10}
}

@article{agarwal2025gpt,
  title={{gpt-oss-120b} \& {gpt-oss-20b} model card},
  author={Agarwal, Sandhini and Ahmad, Lama and Ai, Jason and Altman, Sam and Applebaum, Andy and Arbus, Edwin and Arora, Rahul K and Bai, Yu and Baker, Bowen and Bao, Haiming and others},
  journal={arXiv preprint arXiv:2508.10925},
  year={2025},
  doi={10.48550/arXiv.2508.10925}
}

@article{team2026gemma,
  title={Gemma 4 technical report},
  author={{Gemma Team}},
  journal={arXiv preprint arXiv:2607.02770},
  year={2026},
  doi={10.48550/arXiv.2607.02770}
}

@article{blakeman2025nemotron,
  title={Nemotron 3 nano: Open, efficient mixture-of-experts hybrid mamba-transformer model for agentic reasoning},
  author={Blakeman, Aaron and Grattafiori, Aaron and Basant, Aarti and Gupta, Abhibha and Khattar, Abhinav and Renduchintala, Adi and Vavre, Aditya and Shukla, Akanksha and Bercovich, Akhiad and Ficek, Aleksander and others},
  journal={arXiv preprint arXiv:2512.20848},
  year={2025},
  doi={10.48550/arXiv.2512.20848}
}

@article{liu2021indoor,
  title={Indoor navigation supported by the Industry Foundation Classes ({IFC}): A survey},
  author={Liu, Liu and Li, Bofeng and Zlatanova, Sisi and van Oosterom, Peter},
  journal={Automation in Construction},
  volume={121},
  pages={103436},
  year={2021},
  publisher={Elsevier},
  doi={10.1016/j.autcon.2020.103436}
}

@article{alattas2021mapping,
  title={Mapping private, common, and exclusive common spaces in buildings from {BIM/IFC} to {LADM}. A case study from Saudi Arabia},
  author={Alattas, Abdullah and Kalogianni, Eftychia and Alzahrani, Thamer and Zlatanova, Sisi and Van Oosterom, Peter},
  journal={Land Use Policy},
  volume={104},
  pages={105355},
  year={2021},
  publisher={Elsevier},
  doi={10.1016/j.landusepol.2021.105355}
}

@article{aleksandrov2024bim,
  title={{BIM}-based connectivity graph and voxels classification for pedestrian-hazard interaction},
  author={Aleksandrov, Mitko and Zlatanova, Sisi and Heslop, David J and Diakite, Abdoulaye},
  journal={Journal of Spatial Science},
  volume={69},
  number={3},
  pages={699--719},
  year={2024},
  publisher={Taylor \& Francis},
  doi={10.1080/14498596.2023.2281923}
}

@article{zhang2026instruction,
  title={Instruction tuning for large language models: A survey},
  author={Zhang, Shengyu and Dong, Linfeng and Li, Xiaoya and Zhang, Sen and Sun, Xiaofei and Wang, Shuhe and Li, Jiwei and Hu, Runyi and Zhang, Tianwei and Wang, Guoyin and others},
  journal={ACM Computing Surveys},
  volume={58},
  number={7},
  pages={1--36},
  year={2026},
  publisher={ACM New York, NY},
  doi={10.1145/3777411}
}

@article{pan2026llm,
  title={{LLM}-enabled multi-agent framework for natural language interaction with graph-based digital twins},
  author={Pan, Yuandong and Wang, Mudan and Lu, Linjun and Lamsal, Rabindra and P{\"a}rn, Erika and Zlatanova, Sisi and Brilakis, Ioannis},
  journal={Automation in Construction},
  volume={183},
  pages={106791},
  year={2026},
  publisher={Elsevier},
  doi={10.1016/j.autcon.2026.106791}
}

@article{team2025gemma,
  title={Gemma 3 technical report},
  author={Team, Gemma and Kamath, Aishwarya and Ferret, Johan and Pathak, Shreya and Vieillard, Nino and Merhej, Ramona and Perrin, Sarah and Matejovicova, Tatiana and Ram{\'e}, Alexandre and Rivi{\`e}re, Morgane and others},
  journal={arXiv preprint arXiv:2503.19786},
  year={2025},
  doi={10.48550/arXiv.2503.19786}
}

@misc{qwenteam2026qwen3627b,
  title        = {{Qwen3.6-27B}: Flagship-Level Coding in a 27B Dense Model},
  author       = {{Qwen Team}},
  year         = {2026},
  month        = apr,
  howpublished = {Qwen Blog},
  url          = {https://qwen.ai/blog?id=qwen3.6-27b},
  note         = {Accessed 2026-08-10}
}

@article{grattafiori2024llama,
  title={The {Llama 3} herd of models},
  author={Grattafiori, Aaron and Dubey, Abhimanyu and Jauhri, Abhinav and Pandey, Abhinav and Kadian, Abhishek and Al-Dahle, Ahmad and Letman, Aiesha and Mathur, Akhil and Schelten, Alan and Vaughan, Alex and others},
  journal={arXiv preprint arXiv:2407.21783},
  year={2024},
  doi={10.48550/arXiv.2407.21783}
}

@inproceedings{robinson2022leveraging,
  title={Leveraging large language models for multiple choice question answering},
  author={Robinson, Joshua and Wingate, David},
  booktitle = {International Conference on Learning Representations},
  year      = {2023},
  url       = {https://openreview.net/forum?id=yKbprarjc5B}
}

@inproceedings{balepur2025these,
  title={Which of these best describes multiple choice evaluation with {LLM}s? a) forced b) flawed c) fixable d) all of the above},
  author={Balepur, Nishant and Rudinger, Rachel and Boyd-Graber, Jordan Lee},
  booktitle={Proceedings of the 63rd Annual Meeting of the Association for Computational Linguistics (Volume 1: Long Papers)},
  pages={3394--3418},
  year={2025},
  doi={10.18653/v1/2025.acl-long.169}
}

@article{bommasani2021opportunities,
  title={On the opportunities and risks of foundation models},
  author={Bommasani, Rishi and Hudson, Drew A and Adeli, Ehsan and Altman, Russ and Arora, Simran and von Arx, Sydney and Bernstein, Michael S and Bohg, Jeannette and Bosselut, Antoine and Brunskill, Emma and others},
  journal={arXiv preprint arXiv:2108.07258},
  year={2021},
  doi={10.48550/arXiv.2108.07258}
}

@article{lewis2020retrieval,
  title={Retrieval-augmented generation for knowledge-intensive {NLP} tasks},
  author={Lewis, Patrick and Perez, Ethan and Piktus, Aleksandra and Petroni, Fabio and Karpukhin, Vladimir and Goyal, Naman and K{\"u}ttler, Heinrich and Lewis, Mike and Yih, Wen-tau and Rockt{\"a}schel, Tim and others},
  journal={Advances in neural information processing systems},
  volume={33},
  pages={9459--9474},
  year={2020},
  doi={10.48550/arXiv.2005.11401}
}

@article{gao2023retrieval,
  title={Retrieval-augmented generation for large language models: A survey},
  author={Gao, Yunfan and Xiong, Yun and Gao, Xinyu and Jia, Kangxiang and Pan, Jinliu and Bi, Yuxi and Dai, Yixin and Sun, Jiawei and Wang, Haofen and Wang, Haofen},
  journal={arXiv preprint arXiv:2312.10997},
  volume={2},
  number={1},
  year={2023},
  doi={10.48550/arXiv.2312.10997}
}

@inproceedings{francis2018cypher,
author = {Francis, Nadime and Green, Alastair and Guagliardo, Paolo and Libkin, Leonid and Lindaaker, Tobias and Marsault, Victor and Plantikow, Stefan and Rydberg, Mats and Selmer, Petra and Taylor, Andr{\'e}s},
title = {Cypher: An Evolving Query Language for Property Graphs},
year = {2018},
isbn = {9781450347037},
doi = {10.1145/3183713.3190657},
booktitle = {Proceedings of the 2018 International Conference on Management of Data},
pages = {1433–1445},
}

\end{document}